\documentclass[conference]{IEEEtran}
\usepackage{times}

\usepackage[square,numbers,sort&compress]{natbib}
\usepackage{multicol}
\usepackage[bookmarks=true]{hyperref}
\usepackage{graphicx}
\usepackage{xspace}
\usepackage{mathtools}
\usepackage[font=small,labelfont=bf]{caption}
\usepackage{stfloats}

\usepackage[dvipsnames]{xcolor}        

\hypersetup{
    colorlinks=true,
    linkcolor=MidnightBlue,
    filecolor=MidnightBlue,      
    urlcolor=MidnightBlue,
    citecolor=MidnightBlue,
}

\usepackage[capitalise, nameinlink]{cleveref}

 \renewcommand{\baselinestretch}{0.98}
 
\newcommand{\acro}{RT-H\xspace}
\newcommand{\skill}{language motion\xspace}
\newcommand{\skills}{language motions\xspace}
\newcommand{\correction}{\skill correction\xspace}
\newcommand{\corrections}{\skill corrections\xspace}
\newcommand{\joint}{\acro-Joint\xspace}
\newcommand{\onehot}{\acro-OneHot\xspace}
\newcommand{\cluster}{\acro-Cluster\xspace}
\newcommand{\humaniv}{\acro + Human Intervention\xspace}
\newcommand{\iv}{\acro-Intervene\xspace}
\newcommand{\iva}{\acro-InterveneAction\xspace}

\newcommand{\Qdiv}{\textbf{Q1}\xspace}

\newcommand{\Qcontext}{\textbf{Q2}\xspace}
\newcommand{\Qcorr}{\textbf{Q3}\xspace}
\newcommand{\Qgen}{\textbf{Q4}\xspace}

\newcommand{\rt}{\emph{Kitchen}\xspace}
\newcommand{\diverse}{\emph{Diverse+Kitchen}\xspace}
\newcommand{\divonly}{\emph{Diverse}\xspace}

\newcommand{\RV}[1]{#1}

\newcommand{\NOTANON}[1]{#1}

\newcommand{\MS}[2]{#1 (#2)}

\usepackage{array}

\newcolumntype{P}[1]{>{\centering\arraybackslash}p{#1}}

\begin{document}

\title{\acro: Action Hierarchies Using Language}


\author{
Suneel Belkhale$^{1,2}$, Tianli Ding$^1$, Ted Xiao$^1$, Pierre Sermanet$^1$, Quan Vuong$^1$, Jonathan Tompson$^1$, \\
Yevgen Chebotar$^{*,1}$, Debidatta Dwibedi$^{*,1}$, Dorsa Sadigh$^{*,1,2}$ \\\\
$^1$Google DeepMind, $^2$Stanford University
}



%

\twocolumn[{%
\renewcommand\twocolumn[1][]{#1}%
\maketitle

\vspace{-10pt}
\begin{center}
    \captionsetup{type=figure}
    \includegraphics[width=\textwidth]{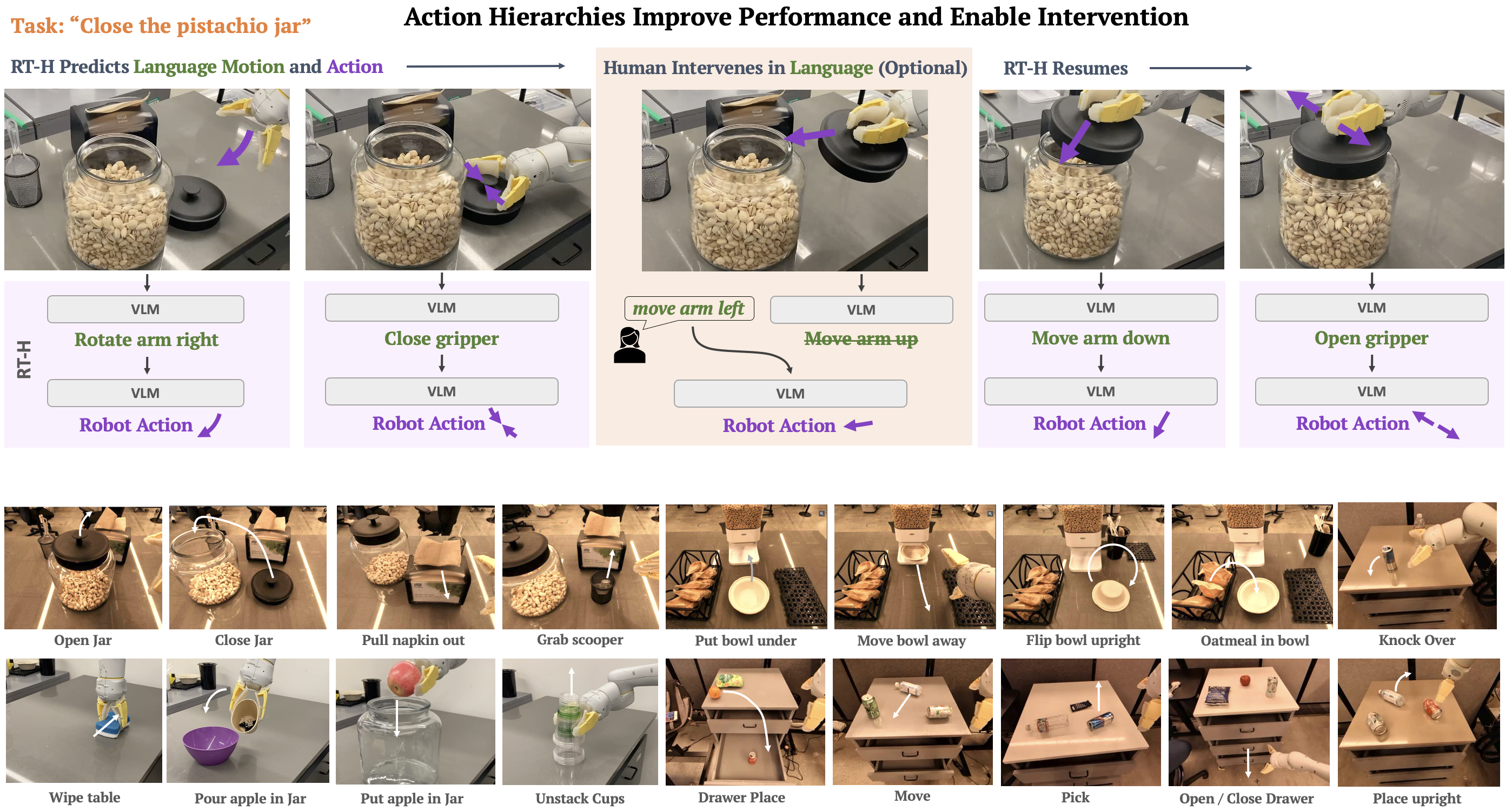}
    \captionof{figure}{Given a task in language like ``close the pistachio jar" and an image of the scene, \acro utilizes a Vision Language Model (VLM) to predict \emph{\skills} like ``move arm forward" and ``rotate arm right", and then conditioned on these \skills, it predicts \emph{actions} for the robot (purple box). This action hierarchy teaches the model the \emph{shared structure} across tasks with many semantically different descriptions (see bottom task examples). These \skills enable better data sharing across diverse multi-task datasets as compared to mapping directly from the task to actions. This hierarchy also enables humans to optionally provide \emph{\corrections} to the robot to prevent task failure, and then to use these new \skills to predict better actions (peach box). Once the human is done intervening, \acro resumes predicting \skills like before.}
    \label{fig:front}
\end{center}
}]

\begin{abstract}
Language provides a way to break down complex concepts into digestible pieces. Recent works in robot imitation learning have proposed learning language-conditioned policies that predict actions given visual observations and the high-level task specified in language. These methods leverage the structure of natural language to share data between semantically similar tasks (e.g., ``pick coke can” and ``pick an apple”) in multi-task datasets. However, as tasks become more semantically diverse (e.g., ``pick coke can" and ``pour cup"), sharing data between tasks becomes harder and thus learning to map high-level tasks to actions requires substantially more demonstration data.
To bridge this divide between tasks and actions, 
our insight is to teach the robot the language of actions, describing low-level motions with more fine-grained phrases like ``move arm forward" or ``close gripper". 
Predicting these \emph{\skills} as an intermediate step between high-level tasks and actions forces the policy to learn the shared structure of low-level motions across seemingly disparate tasks.
Furthermore, a policy that is conditioned on \skills can easily be \emph{corrected} during execution through human-specified \skills. This enables a new paradigm for flexible policies that can learn from human intervention in language. Our method \acro builds an \emph{action hierarchy} using \skills: it first learns to predict \skills, and conditioned on this along with the high-level task, it then predicts actions, using visual context at all stages. Experimentally we show that \acro leverages this language-action hierarchy to learn policies that are more robust and flexible by effectively tapping into multi-task datasets. We show that these policies not only allow for responding to language interventions, but can also learn from such interventions and outperform methods that learn from teleoperated interventions. 
\NOTANON{Our website and videos are found at \href{https://rt-hierarchy.github.io}{rt-hierarchy.github.io}.}
\end{abstract}

\IEEEpeerreviewmaketitle

\section{Introduction}
\label{sec:intro}
Language is the engine of human reasoning, empowering us to break complex concepts into simpler ones, to correct our misunderstandings, and to generalize concepts in new settings. 
In recent years, robots too have begun to leverage language's efficient, compositional structure for breaking down high-level concepts~\cite{brohan2023can}, providing language corrections~\cite{cui2023lilac, Sharma2022CorrectingRP}, or enabling generalization to new settings~\cite{rt22023arxiv}. 
These works often share a common paradigm: given a high-level \emph{task} described in language like ``pick coke can", they learn policies that map observations and task descriptions in language to low-level robot \emph{actions} across large multi-task datasets.
The advantage of language in these settings is to encode the shared structure between similar tasks (e.g., ``pick coke can" vs. ``pick an apple"), reducing the data needed to learn the mapping from tasks to actions.
However as tasks become more diverse, so too does the language describing each task (e.g., ``pick coke can" vs. ``pour a cup"), making it harder to learn the shared structure between different tasks from only the high-level language.

To learn diverse tasks, our aim is to better capture the similarities between these tasks. We observe that language is capable of expressing much more than just the high-level task: we can also express \emph{how} to do the task -- a more fine-grained representation that lies closer to the low-level actions. For example, we can decompose the ``pick coke can" task into a sequence of fine-grained behaviors, which we denote as \emph{\skills}: ``move arm forward", then ``grasp the can", and then ``move the arm up". Our key insight is to leverage \skills as an intermediate prediction layer between high-level task descriptions and low-level actions -- thus building an \emph{action hierarchy} via language motions. Creating such an action hierarchy leads to several benefits:
(1) It enables much better data sharing between different tasks at the level of \skills, leading to better \skill composition and generalization in diverse multi-task datasets. 
For example, even though ``pour a cup" and ``pick up a coke can" are semantically different, they entirely overlap at the \skill level until the object is picked.
(2) Language motions are not merely fixed primitives, but rather learned in the \emph{context} of the current task and scene using the instruction and visual observation. For example, ``move arm forward" alone does not convey how fast to move or in what exact direction vector; that depends on the task and the observation. The contextuality and flexibility of learned \skills introduce a new set of capabilities: it allows humans to provide their own \emph{corrections} to \skills when the policy is not 100\% successful (see center orange box in \cref{fig:front}). Further, the robot can even learn from these human corrections, entirely in the realm of \skills. For example, with ``pick coke can", if the robot closes its gripper early, we can instead tell it to ``move arm forward" for longer, which \acro interprets in context of the current scene. This slight change in \skills is not only easy for a human to provide, but also much easier to learn from compared to correcting individual robot actions.


Motivated by the benefits of \skills, we propose an end-to-end framework, \acro (Robot Transformer with Action Hierarchies), for learning these action hierarchies: at each step, \acro conditions on the observation and the high-level task description to predict the current \skill (\skill query), enabling the model to reason about how to do the task at a fine-grained level. Then \acro uses the observation, the task, and the inferred \skill to predict the action for that step (action query), where the \skill provides additional context to improve the prediction of precise actions  (see purple box in \cref{fig:front}). To extract \skills from our data, we develop an automated approach to extract a simplified set of \skills from robot proprioception, yielding a rich library of over 2500 \skills without any manual annotation effort. We base our model architecture on RT-2, a large Vision-Language Model (VLM) which co-trains on internet-scale vision and language data to improve policy learning~\cite{rt22023arxiv}. \acro uses a single model for both \skill and action queries to leverage this broad internet-scale knowledge for all levels of the action hierarchy.



Experimentally, we find that using a \skill hierarchy yields substantial improvements when ingesting diverse multi-task datasets, outperforming RT-2 by 15\% on a wide range of tasks. We also find that correcting \skills reaches near perfect success rates on the same tasks, demonstrating the flexibility and contextuality of learned \skills. Additionally, fine-tuning our model with \skill interventions outperforms state-of-the-art interactive imitation learning methods such as IWR~\cite{mandlekar2020iwr} by 50\%. Finally, we show that \skills in \acro generalize to variations in scene and objects better than RT-2.

\section{Related Work}
\label{sec:rw}
\noindent In this section, we discuss the role of language in policy learning, how hierarchy has been used in imitation learning, and previous approaches for providing and learning from human corrections on robot policies.

\smallskip \noindent \textbf{Language-Conditioned Policies.}
 In recent years, language has emerged as a powerful goal representation for robotic tasks. In imitation learning (IL), many approaches encode tasks described in language into embeddings using pretrained language models, which are then inputted to a policy that is trained on multi-task robot datasets~\cite{brohan2022rt1, jang2021bcz, stepputtis2020langil, shridhar2021cliport, mees2022matters, sundaresan2023kite}. These pretrained language embeddings lack any visual understanding, so other works jointly train visual and language representations from the ground up, often using large internet-scale datasets and sometimes including robot data~\cite{radford2021clip, nair2023r3m, karamcheti2023voltron, ma2022vip}. The resulting goal representations can then be inputted into a policy to provide both visual and semantic context. More recently, policies built on vision language model (VLM) backbones have become capable of learning actions directly from visual observations and language without the need for pretrained embeddings~\cite{driess2023palme, rt22023arxiv}. All these approaches leverage language to represent the high-level task and often directly predict low-level actions -- but as both language task descriptions become semantically diverse, sharing data between different tasks becomes challenging, so significantly more data is required.

\smallskip \noindent \textbf{Hierarchical Action Representation in Imitation Learning.}
 An alternative approach to boost performance is to impose structure on the multi-task learning problem through the use of \emph{hierarchical action representations}. Several works have explored learning general ``skill" representations as parameterized primitives~\cite{konidaris2012cst, niekum2012unstructured} or embeddings to describe short sequences of actions or interactions with objects, often from diverse multi-task datasets~\cite{krishnan2017ddco, shankar2020learning, kipf2019compile, Shankar2020Discovering, tanneberg2021skid, zhu2022bottom, hakhamaneshi2021hierarchical,wang2017robust, mees2022matters, lynch2020lfp, belkhale2022plato}. While they generally improve performance, they are often quite computationally complex and sensitive to hyperparameters. Another line of work shows the benefits of separating coarse and fine action abstractions in IL,
but requiring both coarse and fine action annotations~\cite{johns2021coarse, belkhale2023hydra}.

Language has also been used to create hierarchy in multi-task learning. When tackling long horizon instructions, many recent approaches use LLMs or VLMs to decompose long horizon instructions into a sequence of tasks specified in language~\cite{brohan2023can,huang2023inner,robovqa2023arxiv,mirchandani2021ella,driess2023palme,hejna2023improving}. Usually, scripted or individually trained policies are used to execute these tasks, limiting scalability. To learn long horizon policies end-to-end, \citeauthor{hu2023ThoughtCloning} train a model to first predict language tasks and then predict actions conditioned on those tasks~\cite{hu2023ThoughtCloning}. This approach is similar to \acro but exists one level higher in the action hierarchy: they do not label or learn from fine-grained \skills. 
A few works explore the usage of more fine-grained language\RV{, for example predefined motion primitives or by predicting object dynamics from verbs~\cite{sharma2022skill, ma2023skill}}. \citeauthor{sharma2022skill} use an LLM to decompose tasks into a sequence of motion primitives~\cite{sharma2022skill}. Yet, these motion primitives are hard-coded and lack the contextuality that is required in more complex settings. \acro learns \skills in the context of both the task and the scene, enabling better policies and more contextual corrections.

\smallskip \noindent \textbf{Interactive Imitation Learning and Correction.}
 Interactive IL methods learn from human feedback during robot execution~\cite{liu2022interactive}. \citeauthor{ross2022dagger} proposed DAgger, which iteratively aggregates expert annotated actions for online rollouts~\cite{ross2022dagger}. While effective, providing such expert annotation is costly, so later works used more selective interventions, for example letting either the human decide when to intervene~\cite{kelly2019hg, mandlekar2020iwr} or letting the policy decide~\cite{hoque2022thriftydagger, hoque2021lazy, zhang2017query, menda2019ensemble}. These methods all require intervention in the action space of the robot, i.e., robot teleoperation~\cite{mandlekar2020iwr} or kinesthetic teaching~\cite{li2021learning, losey2022physical}, which can be challenging for non-experts and hard to scale.

To make intervention more intuitive and scalable, several works have studied language as an intervention medium, for example intervening on incorrect task predictions with human guidance~\cite{robovqa2023arxiv, zha2023distilling}. Correcting language at a more fine-grained level is challenging. Several works define a fixed set of fine-grained language corrections and then map natural language utterances to these correction primitives~\cite{broad2017real}. Later work removes these brittle primitives and replaces them with composable cost functions, but they assume privileged environment knowledge~\cite{Sharma2022CorrectingRP}. Data driven approaches have also been explored, requiring large datasets of language corrections~\cite{bucker2022reshaping,bucker2022latte,co2018guiding,lynch2023interactive} or shared autonomy during deployment~\cite{cui2023lilac}. \acro instead learns to \emph{predict} \skills end-to-end with actions, enabling not just correction in the space of \skills but also efficient learning from those corrections.

\section{\acro: Action Hierarchies using Language}
\label{sec:method}

\begin{figure*}[h!]
    \centering
    \includegraphics[width=\linewidth]{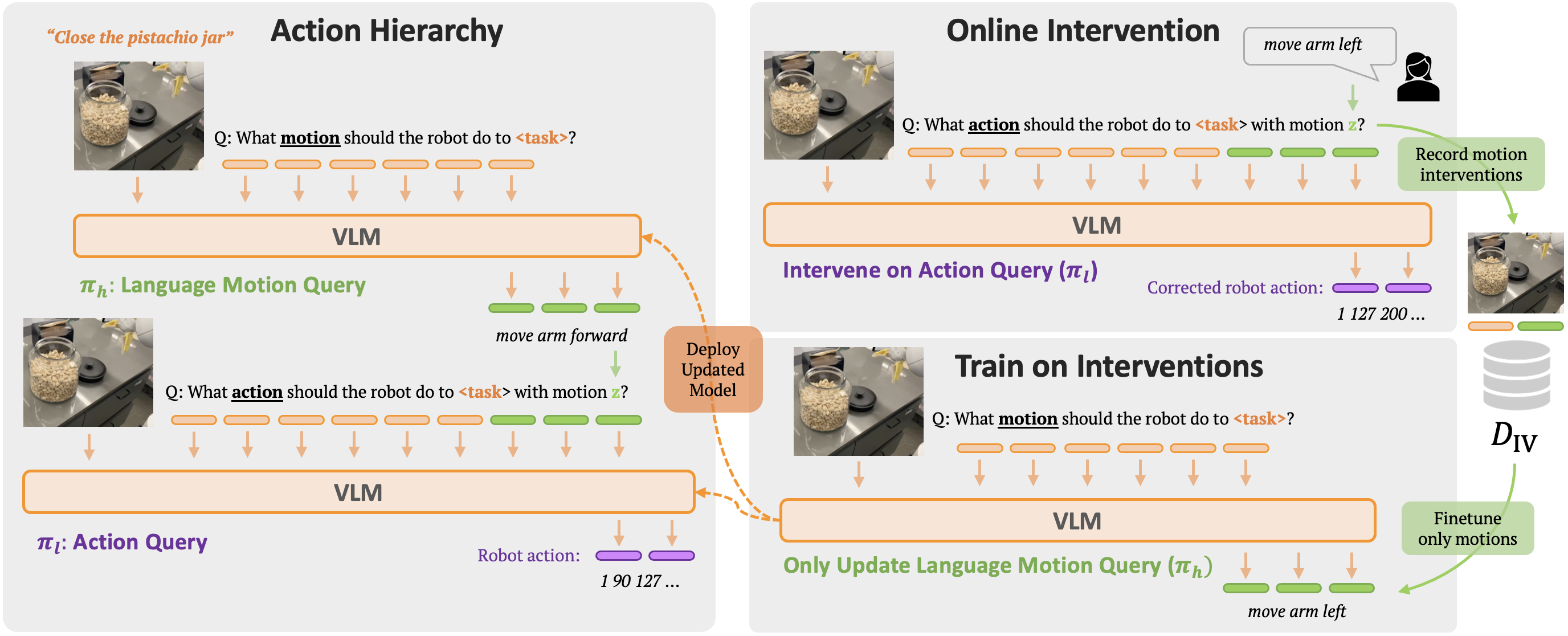}
    \caption{\acro Overview. \textbf{Left:} Our method leverages language to create an action hierarchy for policy learning. We separate the action prediction problem into a \skill query ($\pi_h$), which predicts a fine-grained language motion like ``move arm forward" using the image tokens and task description tokens, and an action query ($\pi_l$), which flexibly decodes this \skill into actions using the context of the task and the scene. We leverage a single VLM for both queries based on RT-2~\cite{rt22023arxiv} that encapsulate the broad prior knowledge in internet-scale data at each level of the action hierarchy. \textbf{Right:} a user can intervene directly on the action query to provide \corrections to robot behavior, for example ``move arm left" instead of ``move arm forward" here (top). To learn from corrections, we can update only the \skill query with the newly labeled \corrections (bottom). Then we deploy the updated model back to the action hierarchy (orange block).}
    \label{fig:method}
    \vspace{-0.35cm}
\end{figure*}

\noindent To effectively capture the shared structure across multi-task datasets -- that is not represented by high-level task descriptions -- our goal is to learn policies that explicitly leverage \emph{action hierarchies}. Specifically, we introduce an intermediate \emph{\skill} prediction layer into policy learning. The \skills describing fine-grained behavior of the robot can capture useful information from multi-task datasets and can lead to performant policies. 

When the learned policies struggle to perform, \skills can again come to rescue: they enable an intuitive interface for online human corrections that are contextual to the given scene. A policy trained with \skills can naturally follow low-level human corrections and successfully achieve the task given the correction data. Additionally, the policy can even be trained on the language correction data and further improve its performance. 


\smallskip \noindent \textbf{\acro Model Overview.}
\acro, shown in \cref{fig:method}, has two key phases:
It first predicts \skills from the task description and visual observation (\emph{\skill query}, top left of \cref{fig:method}), and then conditions on the predicted \skill, the task, and the observation to infer the precise actions (\emph{action query}, bottom left of \cref{fig:method}).
We instantiate \acro using a VLM backbone and following the training procedure from RT-2~\cite{rt22023arxiv}. Similar to RT-2, we leverage the immense prior knowledge in natural language and image processing in internet-scale data through co-training. To incorporate this prior knowledge into all levels of the action hierarchy, a single model learns both the \skill and action queries. 

\subsection{Formalizing Action Hierarchies}
\noindent We are given a dataset $\mathcal{D}=\{(\tau_1, g_1),\dots,(\tau_N,g_N)\}$ of $N$ expert demonstrations ($\tau$) paired with task descriptions in natural language ($g \in \mathcal{G}$), where each $g$ describes exactly one task from a set of $m$ high-level tasks $\{T_i\}_{i=1}^m$. Each demonstration $\tau_i$ consists of a sequence of observations and \emph{action hierarchies} of length $L_i$.
We define an action hierarchy to consist of an intermediate action representation specified in natural language $z \in \mathcal{Z}$, and the low-level action $a \in \mathcal{A}$. Here, the intermediate action is more fine-grained than the high-level task, but more coarse-grained than the low-level action. Thus we write $\tau_i = \{(o_1,z, a_1),\dots,(o_{L_i},z_{L_i},a_{L_i})\}$, with observations $o \in \mathcal{O}$. Our goal is to learn a sequence of policies: a high-level policy $\pi_h: \mathcal{O} \times \mathcal{G} \to \mathcal{Z}$ which maps observations and task descriptions to intermediate actions, and a low-level policy $\pi_l: \mathcal{O} \times \mathcal{G} \times \mathcal{Z} \to \mathcal{A}$ which maps observations, task descriptions, and the intermediate action to the low-level action. Then we define the action hierarchy policy as the composition of these two policies: $\pi(a,z|o,g) = \pi_h(z|o,g) \pi_l(a|o,g,z)$.

In this work, we model the intermediate action representation $z$ using \skills like ``move arm forward" or ``rotate arm right". Note that an action hierarchy can easily be extended to more than just a single level (i.e., $z^1\dots z^K$, in order of how fine-grained they are).



\subsection{\acro: Model and Training Details}
\noindent To model this action hierarchy and acquire the benefits of \skills, our method \acro, shown in \cref{fig:method}, learns $\pi_h$ and $\pi_l$ using a single VLM co-trained with internet-scale data. We instantiate this VLM with the same PaLI-X 55B~\cite{chen2023palix} architecture as RT-2~\cite{rt22023arxiv} -- \acro uses a ViT encoder model to process images into tokens, and then uses an Encoder-Decoder transformer to convert streams of image and natural language tokens into action tokens. These action tokens are produced in the same fashion as RT-2, by discretizing each action dimension into 256 bins and encoding these bins as integer values. Each action $a$ is comprised of delta positions of the end effector, delta axis-angle rotations of the end effector, actions to close or open the gripper, and a termination flag.
\acro constructs two queries to the VLM. First, a \textbf{\skill query} models $\pi_h$, mapping tasks described in language $g$ and the image observations $o$ to \skills $z$ (Encoder sees $g$ and $o$, Decoder predicts $z$). This first stage teaches \acro to first predict correct behavior (\skill) in a coarser and more compressed action space than the low-level robot actions, enabling better modeling of the structure of each task and thus better sharing of sub-trajectories across diverse tasks. Second, the \textbf{action query} models $\pi_l$, mapping the image $o$, task $g$, and the \skill $z$ to action tokens, which then get detokenized into robot actions $a$ (Encoder sees $g$, $o$, and $z$, Decoder predicts $a$). This second stage teaches \acro to be contextual (both with the scene and the task description) in how it decodes \skill $z$ into precise actions to be executed. This extra context is often critical to complete the task successfully, and it is also important for performing and learning from correction, as we discuss in~\cref{sec:method:intervention}. Compared to training both \skill and action autoregressively in one query, using two queries enables (1) specialized prompts for each query, and (2) the \skill $z$ is passed into the Transformer Encoder rather than the Decoder when predicting actions.

\acro is then co-trained using the same PaLI-X~\cite{chen2023palix} training mixture that is used in RT-2, starting from a pre-trained checkpoint. The ViT encoder is frozen for this co-training. \acro replaces the action prediction query in RT-2 with the \skill and action queries at equal sampling rates. Using a single model simplifies the training process, and enables both \skill and action queries to benefit from the broad prior knowledge in the PaLI-X training mixture. See \cref{app:method_details} for model and training details.

\subsection{Extracting Language Motions}
\label{sec:method:labeling}
\noindent While in principle, humans can label the full spectrum of fine-grained \skills, we found that having humans provide these labels offline leads to language inconsistency across the dataset and even inaccuracy in the labeled skills. For example, humans would often mislabel the transitions between skills, or misjudge the direction of motion of the robot due to camera angles. Thus to cheaply extract reliable \skills $z$ at each time step in each episode, we develop an automated labeling scheme relying on robot proprioception information. First, we connect each dimension of the change in robot end effector pose to a spatial dimension (e.g., the z-axis of the position change maps to up and down). Doing this for all 9 action dimensions (3 dimensions for delta position, 3 dimensions for delta orientation, 2 dimensions for base movement, 1 dimension for gripper) we determine a list of the current \emph{dominant} spatial movements of the robot, for example ``move arm up and right", ``close gripper", ``rotate arm counterclockwise", or ``turn base left". Then, we can filter out the dimensions that are below a chosen ``small action" threshold, and then compose the resulting actions in order of the action magnitude. For example, if the robot is predominantly moving the arm forward but also beginning to close gripper, we would extract "move arm forward and close gripper." In this manner, the combinatorial nature of language enables over 2500 \skills to be extracted from a simple set of known motions. Furthermore, since these \skills are derived directly from actions, they hold high predictive power for the actions themselves when running the action query in \acro. Importantly, we fix the details of this procedure for all our experiments and datasets irrespective of the task, and so designing this procedure is a one-time fixed cost for the developer.

Of course, this procedure represents one simple way to define and extract \skills, and many others could exist. For example, one can imagine developing a higher-level object-referential \skill space, for example ``reach the object" or ``grasp the object handle", but this likely requires human annotation or robust object detection and tracking. One could also label at an even more fine-grained level, for example describing the rate of motion like ``move arm forward slowly." However, these examples highlight a fundamental trade-off that exists in the chosen abstraction level for \skills: the more fine-grained they are, the harder they would be to predict for the \skill query, but the more guidance they provide to the action query, and vice versa. As we show in \cref{sec:exp}, our choice of \skills strikes a good balance in both \skill and action query accuracy, while also being cheap to label.

\section{RT-H: Inference \& Correction}
\label{sec:method:inf}
\noindent At test time, \acro first runs the \skill query to infer the skill, and then uses this inferred skill in the action query to compute the action. However, this process doubles inference time since the two queries must be run sequentially at each time step. While not a problem for smaller models, with larger model sizes such as the 55B model used in \acro, we will experience unavoidable querying lag. To handle this challenge, we discuss two modes of \skill inference: (1) \textbf{asynchronous querying}: we train just the \skill query in \acro to predict the skill one step into the future. Then at test time, we query the action using the inferred \skill of the \emph{previous} time step, while also predicting the \skill for the next time step. This enables us to batch the queries and thus achieve nearly identical querying lag as RT-2. (2) \textbf{fixed frequency}: we can evaluate skill queries once every $H$ steps, also reducing the amortized lag. In our experiments we opt for asynchronous querying, since skills often need to change at precise time steps that may not align with fixed frequencies.

\subsection{Correction via Language Motions}
\label{sec:method:intervention}
\noindent 
Even when an \acro policy encounters new manipulation settings or fail at a given task, the action hierarchy in \acro makes it possible to \emph{correct} the policy: users can directly intervene on the learned \skills. While the policy might struggle at performing the high-level task, following the lower-level \skills is much easier. 

Intervening on the model is simple in \acro (see top right in \cref{fig:method}), and since it is text-based, all you need is a keyboard or a microphone. Similar to prior interactive imitation learning approaches such as Intervention Weighted Regression (IWR)~\cite{mandlekar2020iwr}, we let the human operator decide when to intervene on the model, e.g., by pressing a key on the keyboard. Once they have entered the correction mode, they can type a new \correction on the keyboard or use hotkeys for common \skills. This new \skill will directly be passed into the action query in \acro (see \cref{fig:method}) to produce a contextual action that aligns with the user's intent. 

We can also display the current predicted \skill for transparency and providing the user additional context, so they know what the robot was planning to do and can better choose their corrections. Then, at a fixed frequency, we requery the user to either enter a new \correction, keep running the previously entered \correction, or exit correction mode. Fixed frequency requerying gives the user time to update their correction or to decide to let the model take over once again.

\smallskip \noindent \textbf{Learning from Correction Data.} To \emph{learn} from these \corrections, we collect the \skill labels that were provided and the associated images for corrections from successful episodes. Then, we do not need to re-train the action query in \acro, since the model already knows how to map the \correction to actions that succeed at the task; instead, we only need to update the \skill query to produce the correct higher level motions (see \cref{fig:method}; bottom right). This significantly reduces the complexity of learning from corrections, since we only need to learn minor changes in the smaller \skill space rather than the large action space. Like IWR~\cite{mandlekar2020iwr}, we co-train with a mixture of the original dataset (both action and \skill queries) and the dataset of corrections (just \skill query). See \cref{app:interventions} for more mixture details. Of course, since we use a single model for both \skill and action queries, updating only one will likely still update the other -- co-training \acro on both queries in the \emph{original dataset} helps maintain action prediction performance while still learning the minor changes in \skill space through the intervention dataset.

\section{Experiments}
\label{sec:exp}
\noindent To comprehensively evaluate the performance of \acro, we study four key experimental questions:
\begin{itemize}
    \item \Qdiv (\textbf{Performance}): Do action hierarchies with language improve policy performance on diverse multi-task datasets?
    \item \Qcontext (\textbf{Contextuality}): Are learned \skills in \acro contextual to the task and scene?
    \item \Qcorr (\textbf{Corrections}): Is training on \corrections better than teleoperated corrections?
    \item \Qgen (\textbf{Generalization}): Do action hierarchies improve robustness to out-of-distribution settings?
\end{itemize}


\begin{figure*}[b]
    \centering
    \includegraphics[width=\linewidth]{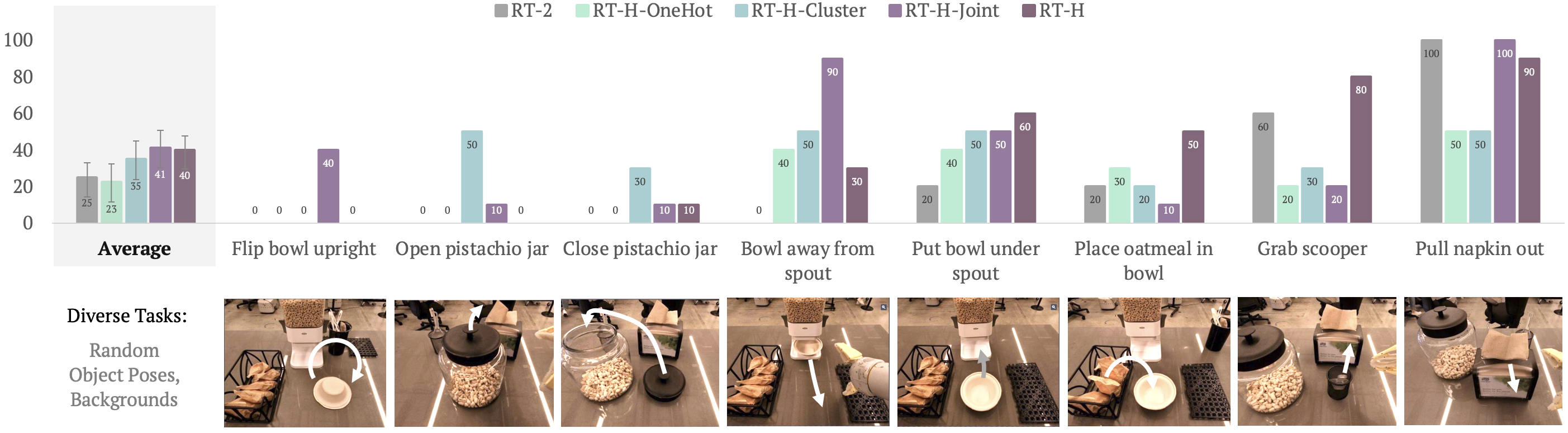}
    \caption{Results on \diverse multi-task dataset, consisting of eight challenging evaluation tasks. \RV{95\% Wilson Score confidence intervals~\cite{wilson1927score} are shown on the average success rates (left).} \acro outperforms RT-2 by 15\% on average, getting higher performance on 6/8 of the tasks. Replacing language with class labels (\onehot) drops performance significantly. Using action clusters via K-Means~\cite{lloyd1982kmeans} instead of the automated motion labeling procedure leads to a minor drop in performance as well (\cluster), demonstrating the utility of \skills as the intermediate action layer. 
    }
    \label{fig:phase4_barplots}
    \vspace{-0.35cm}
\end{figure*}

\noindent \textbf{Dataset}: We utilize a large multi-task dataset consisting of 100K demonstrations with randomized object poses and backgrounds. This dataset combines the following datasets:
\begin{itemize}
    \item \rt: The dataset used in RT-1~\cite{brohan2022rt1} and RT-2~\cite{rt22023arxiv}, consisting of 6 semantic task categories in 70K demonstrations.
    \item \divonly: A new dataset consisting of more complex range of tasks, with over 24 semantic task categories, but just 30K demonstrations (see \cref{app:datasets} for more details).
\end{itemize}
We call this combined dataset the \diverse (\emph{D+K}) dataset, and it is labeled with \skills using our automated procedure described in \cref{sec:method:labeling}. We evaluate our method trained on the full \diverse dataset on eight tasks that are a representative sample of its hardest tasks:
\begin{enumerate}
    \item ``flip bowl upright on the counter"
    \item ``open pistachio jar"
    \item ``close pistachio jar"
    \item ``move bowl away from cereal dispenser"
    \item ``put bowl under cereal dispenser"
    \item ``place oatmeal packet in the bowl"
    \item ``grab scooper from basket"
    \item ``pull napkin from dispenser"
\end{enumerate}
These eight tasks, shown in \cref{fig:phase4_barplots}, were chosen because they require complex sequences of motions and high precision.

\smallskip
\noindent \textbf{Methods}: 
We study and compare the following methods, including ablating a number of choices in \acro:
\begin{itemize}
    \item \textbf{\acro} is our proposed method in this work, and we use the asynchronous querying variant for these experiments (see \cref{sec:method:inf}).
    \item  \textbf{\joint} is also our method but using a single autoregressive query to produce both \skill and action, rather than querying the VLM twice with two different prompts for each query. \joint first outputs \skill then action (where action is still conditioned on \skill). While both \acro and \joint are autoregressive on the \skill, \acro uses distinct queries for action and \skill (``What \emph{motion} \dots" vs. ``What \emph{action} \dots, given motion \dots"), whereas \joint has just one query (``What \emph{motion and action} \dots"). More specifically, \acro passes the \skill to the Encoder in the action query, while \joint treats the \skill as a Decoder input when predicting the action. Thus, we expect this to perform comparably to \acro.
    \item \textbf{\cluster} is an ablation of the automated \skill labeling procedure, which instead clusters actions directly using K-means~\cite{lloyd1982kmeans} into a set of classes with integer labels. These class labels are used in place of the automatically labeled \skills in \acro.
    \item \textbf{\onehot} ablates the use of language to represent motions in \acro by replacing each unique \skill with an integer class label.
    \item \textbf{RT-2} is a flat model that does not use any action hierarchy~\cite{rt22023arxiv}.
    \item \textbf{\humaniv} involves having a human correct only the \skills during execution, but still using the action query from \acro (top right of \cref{fig:method}). \humaniv is a variant of our method that enables humans to intervene so we expect it to be an upperbound for the other non-intervention-based methods. We discuss this in \cref{sec:exp:intervention}.
    \item \textbf{\iv} is an extension of \acro method additionally trained on human intervention data using \corrections. We discuss this in \cref{sec:exp:intervention}.
    \item \RV{\textbf{\iva} is an ablation of \iv method that trains on \emph{both} the action corrections and \corrections from human intervention data. We discuss this in \cref{sec:exp:intervention}.}
    \item \textbf{RT-2-IWR} is the interactive version of RT-2, which is additionally trained with human interventions in the form of teleoperated demonstrations -- in contrast to \corrections\ -- and is compared to \iv in \cref{sec:exp:intervention}.
\end{itemize}
Note that \joint, \cluster, and \onehot are variants of \acro that still utilize an action hierarchy.  See \cref{app:method_details} for exact queries and a deeper dive into each \acro variant implementation.


In \cref{sec:exp:diverse}, we first train and evaluate the performance of \acro on a diverse multi-task dataset (\Qdiv). In \cref{sec:exp:contextual}, we qualitatively analyze the learned \skills across various tasks to see how \skills adapt to the context (\Qcontext). In \cref{sec:exp:intervention}, we collect and train on \corrections on top of \acro, demonstrating that training on \corrections improves policy performance (\Qcorr). Finally, in \cref{sec:exp:generalization} we test the robustness of \acro to variations in scenes, objects, and tasks (\Qgen).

\subsection{\acro on Diverse Multi-Task Datasets}
\label{sec:exp:diverse}
Here, we discuss how action hierarchies can improve policy performance addressing \Qdiv. We will first discuss the online performance of \acro and its variants when trained on \diverse dataset. We then present offline performance metrics to further analyze the role of \skills.

\smallskip
\noindent \textbf{On-Robot Performance}: \cref{fig:phase4_barplots} illustrates the performance of each method when trained on the \diverse dataset and evaluated on the 8 selected tasks within this dataset discussed earlier. Checkpoints are chosen using validation action MSE, and then run for 10 controlled trials for each task (80 total trials per method). \acro outperforms RT-2 on most of the tasks, \textbf{surpassing RT-2 by 15\% on average}, which strongly supports the benefit of action hierarchies (\Qdiv), despite using no additional human annotation. See \cref{app:more_results} for the success rates for each stage of each task, where we see that \acro makes more progress towards success in 7/8 tasks. Furthermore, whereas RT-2 achieves nonzero performance on only 4/8 tasks, \acro is nonzero on 6/8 tasks and \joint is nonzero on all the tasks, suggesting that \acro and \joint are better at managing the diversity of tasks in the dataset.

\smallskip
\noindent \textbf{Ablations}: \joint does comparably to \acro, showing that \acro is robust to the exact querying mechanism. 
\cluster replaces the automating labeling procedure with action clustering, and without language it performs slightly worse than \acro on average. Interestingly, \cluster does better on the hardest tasks in the evaluation set (open and close pistachio jar). We hypothesize that since \cluster uses clusters derived from the dataset, its clusters provide even more fine-grained action context than our labeling procedure, allowing it to outperform \acro in precise tasks; however, the lack of language makes predicting clusters harder than predicting \skills when using broad datasets, leading to worse performance for \cluster on the broader set of tasks. \onehot replaces \skills with onehot class labels, and it performs much worse than \acro despite being derived from the same underlying \skills. Thus, while action hierarchy itself gets us part of the way, the structure of language greatly improves \skill and action prediction.

\smallskip
\noindent \textbf{Offline Performance}:
We investigate if \skills as an intermediate layer for action prediction has any noticeable effect by comparing the offline validation mean squared error (MSE) for end-to-end action prediction across \acro and its joint variant \joint vs. the flat RT-2 model (\Qdiv). The end-to-end MSE reflects how well each model learns action prediction. For \acro, we also study the action validation MSE when using the \emph{ground truth} (GT) \skill that was labeled in the data as input to the action query ($\pi_l$), rather than inputting the inferred \skill from the \skill query ($\pi_h$). This ground truth MSE reflects how informative the true \skill is for predicting the actions.
In \cref{tab:mse}, we report the minimum MSE across training checkpoints for \acro, \joint, and RT-2 when trained on either the \diverse dataset or the \rt dataset. \acro has roughly a \textbf{20\% lower MSE than RT-2}, and \joint has a \textbf{5-10\% lower MSE than RT-2}, demonstrating that action hierarchies help improve action prediction offline in large multi-task datasets. Using two queries (\acro) instead of one (\joint) also seems to improve action prediction, which could stem from how the \skill gets passed into the model (through the encoder for \acro vs. through the decoder for \joint). \acro (GT) uses the ground truth MSE metric, and we find the gap with the end to end MSE is 40\%, illustrating that the correct labeled \skills are highly informative for predicting actions.

\begin{table}[h]
\centering
\begin{tabular}{P{1.2cm}P{1.2cm}ccc|c}
        Train Dataset & Eval Dataset  & RT-2 & RT-H-Joint & RT-H &  RT-H (GT) \\ \hline
\rt      & \rt & 30.2          & 28.22         & \textbf{24.9}  & 17.9        \\
\emph{D+K} & \divonly & 27.7      & 25.44        & \textbf{23.6}  & 17.8  
\end{tabular}
\captionof{table}{Best checkpoint Mean Squared Error (MSE) for end-to-end action prediction on the validation set for models (columns) trained on different multi-task datasets (rows). \rt refers to the data used to train RT-1~\cite{brohan2022rt1} and RT-2~\cite{rt22023arxiv} (70K demonstrations), \diverse (\emph{D+K}) refers to a combination of \rt and the more complex set of tasks (30K demonstrations). We also report the MSE of using the \emph{ground truth} \skill (the labeled \skill) for the action query in \acro (GT) rather than the inferred \skill from the \skill query. \acro and \joint achieve lower MSE on both datasets compared to RT-2, illustrating the benefits of action hierarchies for ingesting multi-task datasets compared to flat models like RT-2. Also, \acro has lower MSE than \joint.}
\label{tab:mse}
\end{table}

\subsection{Contextual \& Flexible Language Motions}
\label{sec:exp:contextual}
\noindent In this section, we analyze (1) \textbf{contextuality}: how well the actions for a single in-distribution \skill adapt to the context of the scene and the task instruction, and (2) \textbf{flexibility}: how well \acro responds to out-of-distribution \skills. 

\begin{figure}[h!]
    \centering
    \includegraphics[width=\linewidth]{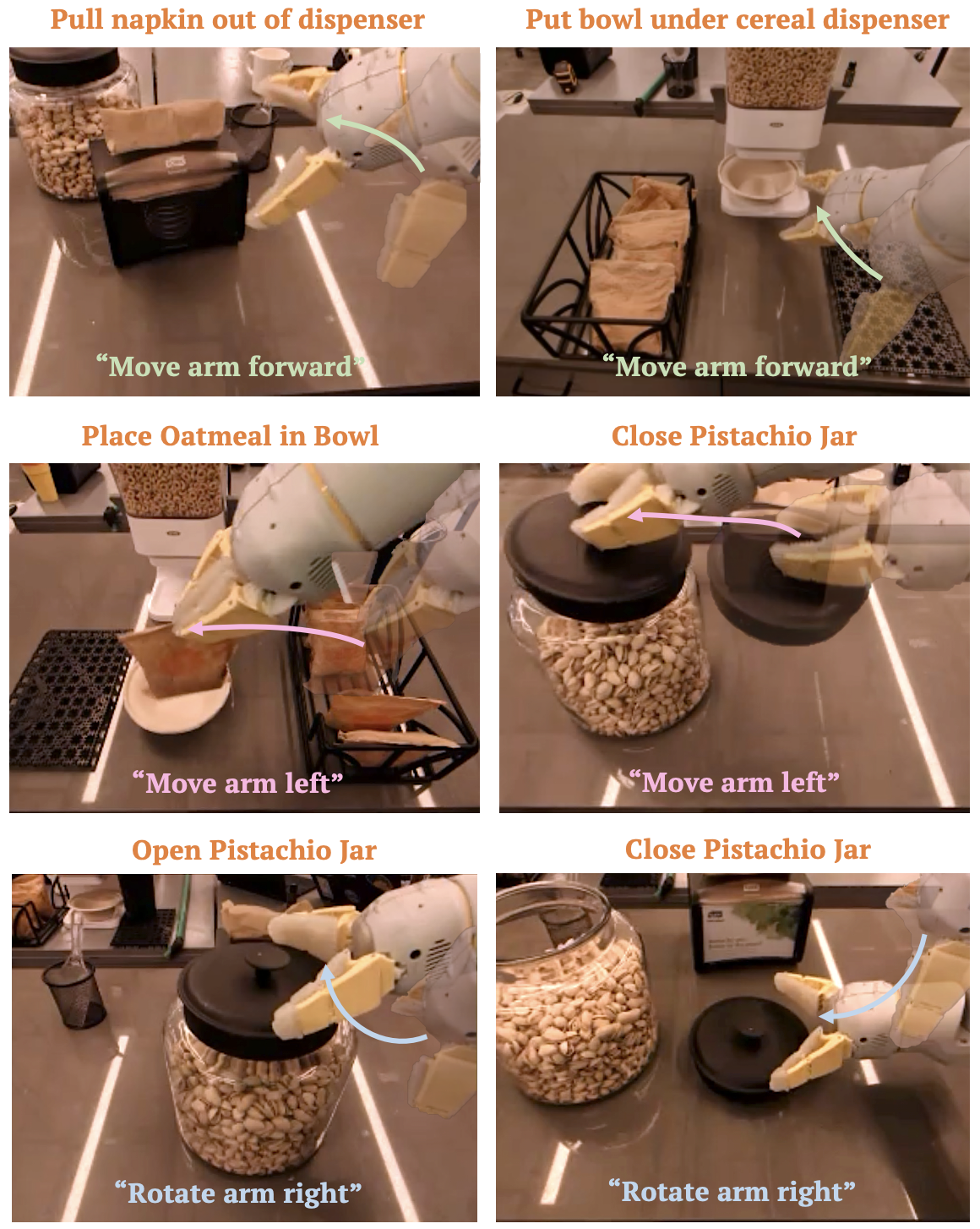}
    \caption{Examples showing how \skills depend on the \emph{context} of the scene and task, taken from online evaluations of \acro trained on the \diverse dataset. For each row, the given \skills (``move arm forward", ``move arm left", ``rotate arm right") manifest with different variations (columns) depending on the task and observation, such as subtle changes in speed, non-dominant axes of movement, e.g., rotation for ``move arm forward", and even gripper positions.}
    \label{fig:context}
    \vspace{-0.35cm}
\end{figure}

\smallskip
\noindent \textbf{Contextuality}: We illustrate several examples of contextual motions taken from online evaluations of \acro in \cref{fig:context}. We see that the same \skills often lead to subtle changes in actions to complete the task, while still respecting the higher level \skill (\Qcontext). For example, for ``move arm forward" in the top left example in \cref{fig:context}, the arm moves generally forward but also towards the object of interest, the napkin dispenser. It also slightly tilts the arm to more easily grasp the napkin. In the top right, we see that the same command leads to a slight downward angle and a rotation of the gripper to avoid colliding with the cereal dispenser. For ``move arm left" in the middle left of \cref{fig:context}, we similarly see that left implies moving the oatmeal packet precisely above the bowl, while in the middle right, left implies precise motion of the lid to latch onto the jar. It would be immensely challenging to design a single ``move arm left" primitive to capture this contextuality. We see a similar behavior for ``rotate arm right" in the bottom row of \cref{fig:context}, where in the left case the arm rotates and stays up to sit on top of the closed jar, while in the right case the arm rotates and moves down to reach the lid on the table. \RV{See \cref{app:more_results} for a quantitative analysis of \skill contextuality.}

\begin{figure}[h!]
    \centering
    \includegraphics[width=\linewidth]{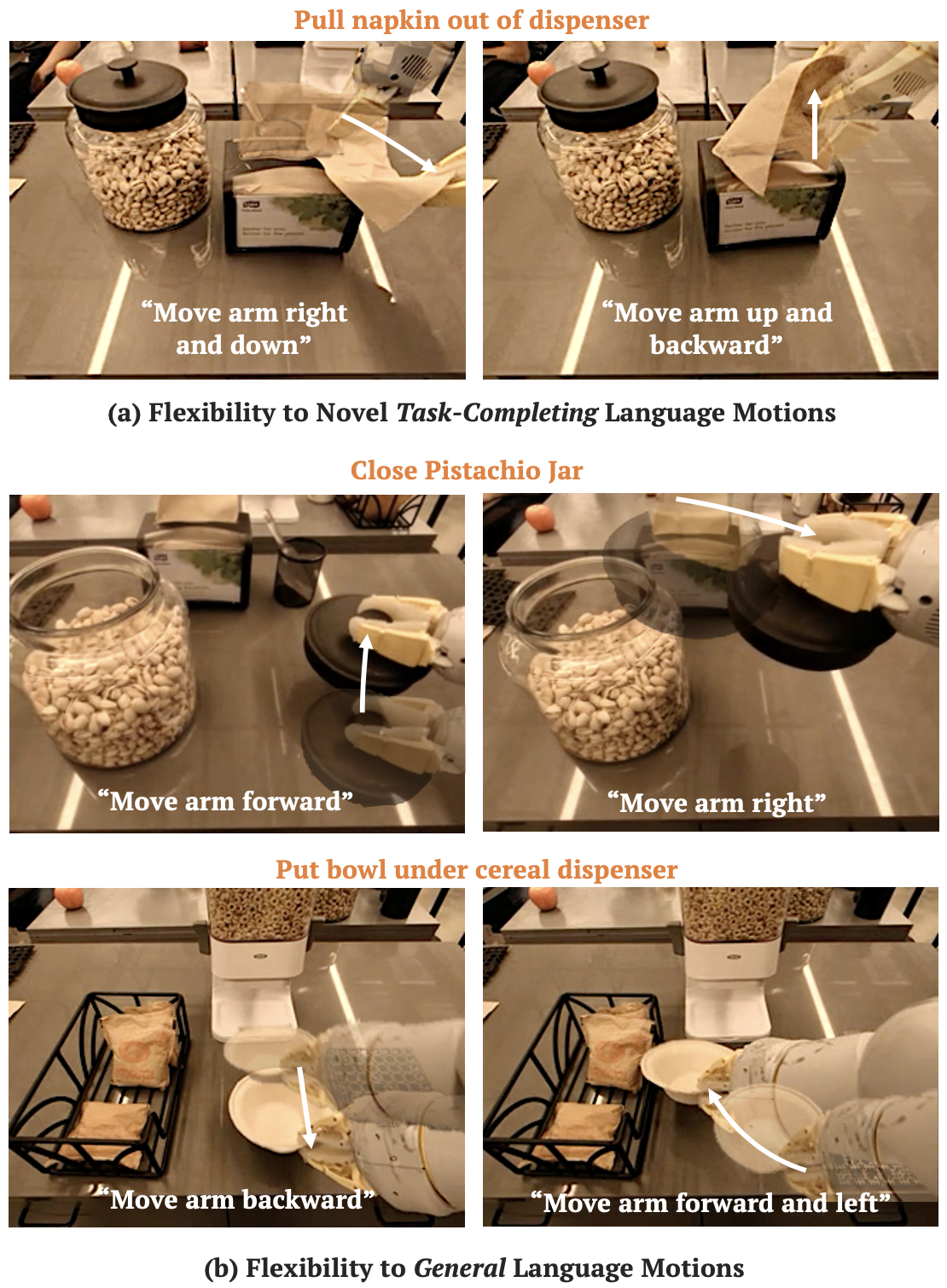}
    \caption{Examples of the flexibility of learned \skills. In the top row (a) we correct \acro using two different task-completing \skills for pulling the napkin out of the dispenser, either ``right and down" or ``up and backward", showing \acro performs both correctly. For the bottom two rows (b), we still correct \skills but ask \acro to perform a more general set of \skills for each task, demonstrating that \acro is often flexible even to completely out-of-distribution \skills for a given task.}
    \label{fig:flex}
    \vspace{-0.35cm}
\end{figure}

\begin{figure*}[h!]
    \centering
    \includegraphics[width=\linewidth]{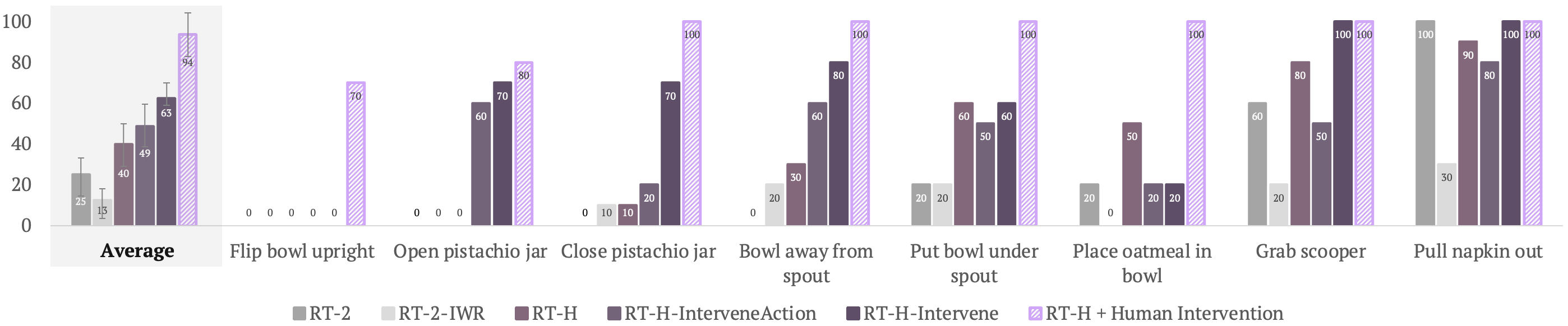}
    \caption{Results for Corrections on models trained on the \diverse multi-task dataset, for the same eight evaluation tasks as in \cref{fig:phase4_barplots}. \RV{95\% Wilson Score confidence intervals~\cite{wilson1927score} are shown on the average success rates (left).} RT-2-IWR is trained on teleoperation corrections from rolling out RT-2, while \iv is trained on \skill corrections from rolling out \acro. \RV{\iva is trained on both \skill and action correction data.} We see \iv both improves upon \acro and substantially outperforms RT-2-IWR, suggesting that \skills are a much more sample efficient space to learn corrections than teleoperated actions. \RV{\iva performs better than \acro, but fine-tuning actions sometimes leads to policy degeneration, since actions produced by \acro during intervention can be suboptimal.}}
    \label{fig:phase4_iv_barplots}
    \vspace{-0.35cm}
\end{figure*}

\smallskip
\noindent \textbf{Flexibility}: In \cref{fig:flex}, we demonstrate the flexibility of \acro by intervening on \skills in \acro online to instead perform out-of-distribution \skills for in-distribution tasks. In the first row (a), \acro is tested with two valid ways of completing the ``pull napkin" task, and we find it responds correctly to both. Despite each of the \skills demonstrated in \cref{fig:flex} being out-of-distribution for the task, \acro is capable of following these new \skills with ease (\Qcontext). In the bottom two rows (b) of \cref{fig:flex}, we find that \acro is also flexible to more general \skills that are not specific to the task and thus not seen in the training data. For example, in the middle right example, moving the arm away from the jar is not a common \skill for ``close pistachio jar", but \acro is still able to act correctly in response to this \skill . Being flexible to more general \skills is critical for responding to a wide variety of \corrections, especially when the task or scene are out-of-distribution and require novel sequences of \skills. 


\smallskip
Overall, we see that \acro is able to maintain the flexibility and contextuality of actions while learning the high-level structure of each task through \skills (\Qcontext). 
\RV{See \cref{app:more_results} for quantitative analysis of contextuality and a qualitative look at \skill multimodality in \acro, along with staged success rates for each method for each task.} 
Next, we leverage these properties to collect and train on \corrections to improve \acro.

\subsection{Training on Online Corrections}
\label{sec:exp:intervention}
\noindent In this section we are interested in how well \acro can learn from \corrections compared to methods without action hierarchy that use teleoperated correction data (\Qcorr). We collect a multi-task \correction dataset and a teleoperated correction dataset for each of the eight tasks in \cref{sec:exp:diverse}. As in prior interactive IL methods~\cite{kelly2019hg, mandlekar2020iwr}, the human decides when to correct in both datasets, usually either anticipating or responding to a task failure. Next we describe the collection and training pipelines for each method.

\smallskip \noindent \textbf{\iv and \iva}: We collect 30 episodes (failed episodes filtered out) of \corrections for each of the eight tasks, using the correction procedure described in \cref{sec:method:intervention}. The base policy used for collection is \acro trained on the \diverse dataset (same as \cref{sec:exp:diverse}). Then, we train \acro on these on-policy corrections in the manner described in \cref{sec:method:intervention} to produce \iv (only training the \skill query on the intervention data). To train \iva, we include both the \skill and action queries when training on the intervention data, at equal sampling rates.

\smallskip \noindent \textbf{RT-2-IWR}: We collect 30 episodes (failed episodes filtered out) of teleoperated corrections for the same eight tasks, using VR-based teleoperation instead of \corrections. Since we only care about learning to correct the failure modes of RT-2, we must use RT-2 trained on the \diverse dataset (same as \iv) as the base policy for collection to ensure fair comparison to \iv. We then train RT-2 on these on-policy corrections using the Intervention Weighted Regression (IWR) method~\cite{mandlekar2020iwr} to produce RT-2-IWR.

Of course, the base policy for RT-2 performs worse than \diverse on these tasks than \acro, so to ensure a fair comparison we focus on the \emph{change} in success rates before and after training on correction data for each method.



\smallskip
\noindent \textbf{Results}: We evaluate both methods in the same eight evaluation tasks as in \cref{sec:exp:diverse}. In \cref{fig:phase4_iv_barplots}, we compare the performance of each method to the pre-correction models, \acro and RT-2, respectively (duplicated from \cref{fig:phase4_barplots}). We also compare \humaniv as an upperbound method that uses online human-in-the-loop \corrections when necessary, but still uses the action query in \acro conditioned on these \corrections. 

First, we see how amenable \acro is to \corrections with \humaniv, which gets very high success rates even for the most precise tasks. This shows that \acro actually does change its behavior in task-relevant ways with \corrections at test time. This further supports the claim in \cref{sec:exp:contextual} that \skills are both flexible and contextual. In addition, this highlights that \skill prediction is often the bottleneck for performance, so we expect that refining \skill prediction through intervention will yield clear improvements.

RT-2-IWR, a state-of-the-art online imitation learning method, sees a degradation in performance from 25\% to 13\% on average, likely due to a combination of relatively small amounts of data per task and the use of only a single round of correction. In addition, we suspect teleoperation-based corrections are more likely to introduce action distributions that are too different from the training data (and thus the base policy). The \corrections in \acro on the other hand are much more consistent with the training data because actions come from the base policy itself (under slight changes in \skill space) and thus easier to learn from.

\iv, on the other hand, substantially outperforms RT-2-IWR in this setting despite using the same amount of data, improving by 60-70\% on the harder precise tasks (open and close pistachio jar). \iv regresses on just one task, ``place oatmeal packet in bowl", where we observed that the robot would successfully grasp the packet very often, but would get stuck predicting ``close gripper". We suspect there is a slight bias towards that \correction in the dataset, and it could be resolved by specifying ``move arm up" as a follow up correction, or by running more rounds of correction. The oatmeal example also highlights how \corrections can make the policy's behavior \emph{interpretable} and thus more intuitive to debug -- more effectively allowing the designer to identify or correct the failure points.

\RV{\iva also improves upon \acro, outperforming it by 9\% on average. We suspect that compared to \iv, \iva suffers from policy degeneration, where new actions in the interventions bias the action distribution toward model generated actions (since we use the action query in \acro to collect \correction data) rather than the true expert distribution. These model generated actions can be sub-optimal and thus can impact performance: for example, we see that close pistachio jar task does not improve as much with \iva as with \iv, because the policy starts producing near-zero actions at states where the intervention-produced actions were small (e.g. after grasping the jar lid).}

Overall, we see that \corrections bring the average success rates of \acro \textbf{from 40\% to 63\% with just 30 episodes of correction} per task. By abstracting actions into the more condensed \skill space, \acro can more quickly learn to improve itself from feedback from \corrections than from teleoperation corrections (\Qcorr).

\smallskip \noindent \RV{
\textbf{Failure Modes}: \acro demonstrates performance boosts on a wide variety of tasks, however the action hierarchy paradigm does lead to interesting failure modes. First, systemic \skill prediction errors can often confuse the action prediction, leading to oscillatory or incorrect behaviors where a flat model might not exhibit similar behaviors. For example, we occasionally noticed the robot would get stuck after closing the gripper by continuing to predict ``close the gripper", likely due to subtle observation changes between ``close the gripper" frames and future frames. Thankfully, these issues are easy to debug and resolve, either through adjusting the automated labeling procedure or through intervention as we have shown.} 

\RV{Second, when collecting \corrections, the human operator is limited by the performance of the underlying action prediction model. When \corrections do not result in the correct behavior, it might become frustrating for the operator. For example if the operator asks the robot to ``move arm left" over the bowl in the middle left row of \cref{fig:context}, but the robot overshoots the bowl, this can make the process slower than teleoperation. This failure mode rarely happens for in-distribution tasks, but as tasks diverge from the data distribution, it becomes more likely.}

\subsection{Generalization}
\label{sec:exp:generalization}
\noindent To evaluate \Qgen, we study three types of generalization: generalization to new scenes (with similar objects but new backgrounds and lighting), to novel objects, and to novel tasks.  We use \acro trained on only the \rt dataset~\cite{brohan2022rt1} unless otherwise noted (i.e., not including the \divonly data), which consists of the following training and evaluation tasks on various objects:
\begin{enumerate}
    \item ``knock over"
    \item ``drawer place"
    \item ``move"   
    \item ``pick"
    \item ``open / close drawer"
    \item ``place upright"
\end{enumerate}

\begin{figure*}
    \centering
    \includegraphics[width=\linewidth]{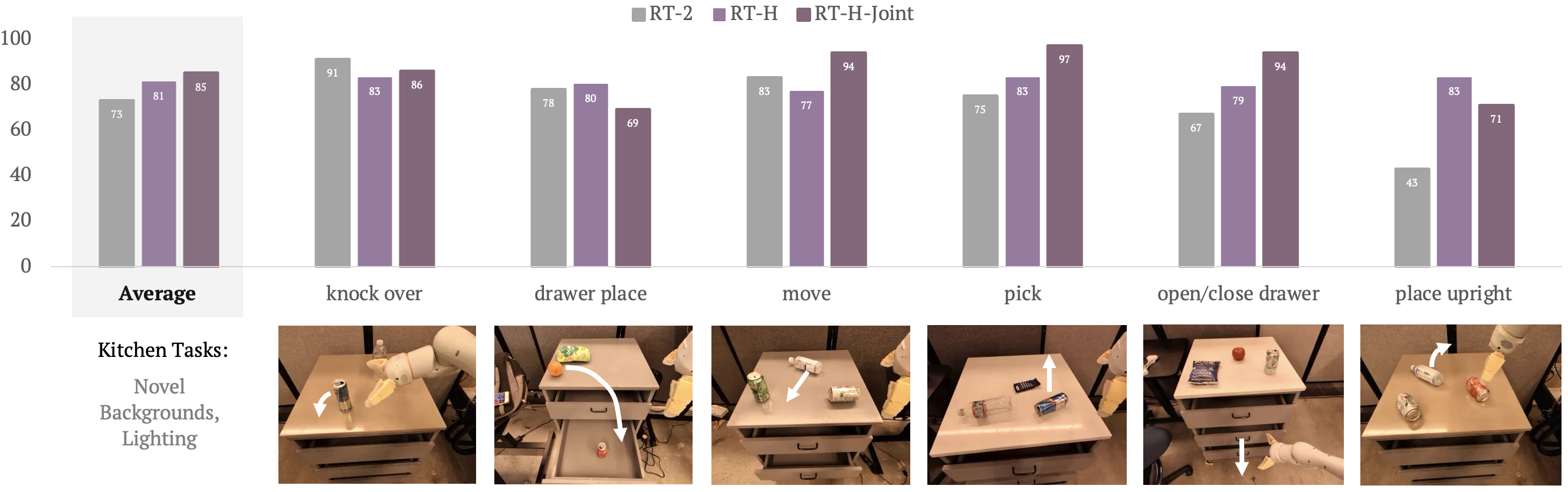}
    \caption{Results when models trained on \rt data~\cite{brohan2022rt1} are deployed on the same tasks, but in a new building with novel backgrounds, lighting, and flooring. \acro and \joint each outperform RT-2, suggesting that the use of action hierarchy helps the policy generalize to novel scenes. RT-2 struggles particularly with placing upright and opening and closing the drawers in these new scenes.}
    \label{fig:generalize}
    \vspace{-0.35cm}
\end{figure*}

\smallskip \noindent \textbf{Generalization to New Scenes}: We evaluate each task in the \rt dataset in new environments, specifically in a new building consisting of varying lighting, and diverse backgrounds and floors. 
In \cref{fig:generalize}, we see that \acro and \joint are more robust to changes in scenes, with especially large deltas for the hardest tasks of ``place upright" task and ``open / close drawer" tasks (\Qgen).

\smallskip \noindent \textbf{Generalization to New Objects}: We evaluate ``pick" and ``move" under object generalization, using 50 evaluations of objects unseen during training such as pears, coconut water, and oreos. In \cref{tab:generalize_objects}, we find that \acro achieves 65\% on these tasks, whereas RT-2 gets 55\%. As shown in \cref{app:more_results}, \acro also progresses farther in each task (in terms of stages of each task) compared to RT-2 on average (\Qgen).

\begin{table}[h]
\centering
\begin{tabular}{l|P{2cm}|P{2cm}|P{2cm} }
& pick & move & \textbf{Average} \\ \hline\hline
RT-2 & 60 & 50 & 55 \\
RT-H & 70 & 60 & 65  \\
\end{tabular}
\captionof{table}{We evaluate RT-2 and \acro trained on \rt data~\cite{brohan2022rt1} on the ``pick" and ``move" tasks but under novel objects for 50 scenarios total. \acro outperforms RT-2, demonstrating that action hierarchy helps the policy generalize to novel objects.}
\label{tab:generalize_objects}
\vspace{-0.3cm}
\end{table}

\begin{figure*}[h!]
    \centering
    \includegraphics[width=\linewidth]{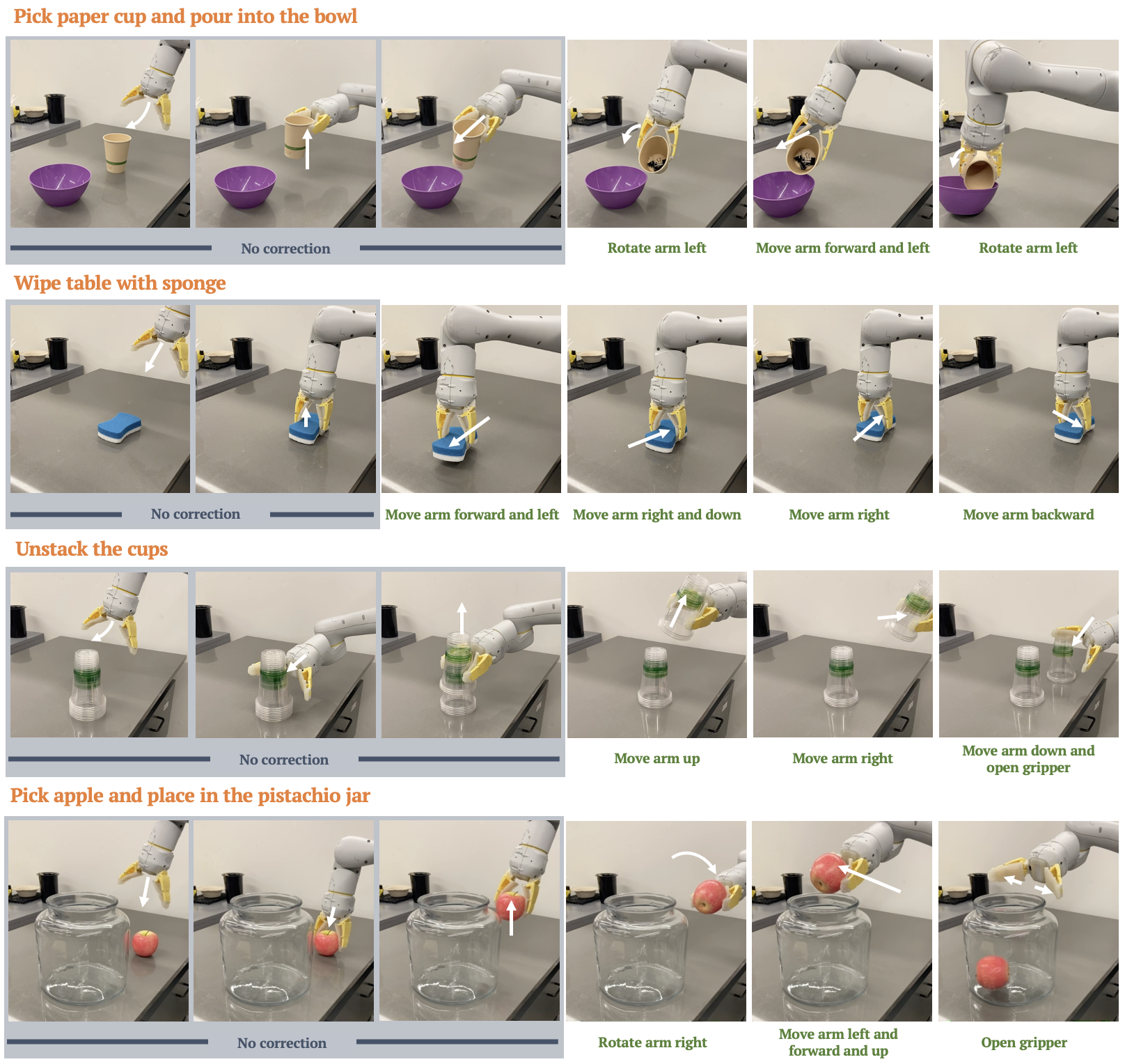}
    \caption{We show the generalization capabilities of \acro with completely unseen tasks with minimal correction. By breaking down tasks into \skills, \acro learns the shared structure between seemingly diverse tasks. This allows it to generalize \skills to new tasks, as shown in the first part of each task, where \acro performs the picking phases easily. We also show that when \acro cannot zero-shot generalize, \corrections often do generalize, allowing it to complete these tasks with just a few well-timed corrections.}
    \label{fig:newtasks}
    \vspace{-0.4cm}
\end{figure*}

\smallskip \noindent \textbf{Generalization to New Tasks with Limited Corrections}: While zero-shot success on out-of-distribution tasks is quite difficult, in \cref{fig:newtasks}, we qualitatively demonstrate that even for unseen tasks, \acro requires just a few well-timed corrections to succeed at the task. For these examples, we use the version of \acro trained on the \diverse dataset to provide \acro with \skills demonstrated in a wide variety of contexts. \cref{fig:newtasks} also shows the shared structure between seemingly diverse tasks: each of these tasks require some picking behavior to begin the task, and by learning the shared structure of \skills across many diverse tasks, \acro can complete the picking stage without any correction (\Qgen). Even when \acro is no longer able to generalize its \skill prediction, we see that \corrections often do generalize, allowing us to successfully complete the task with just a few corrections (\Qcontext, \Qgen). This demonstrates the potential of \skills for scaling up data collection for novel tasks.


\section{Conclusion} 
\label{sec:conclusion}
In this work, we introduce \acro, which leverages \skills like ``move arm forward" as an intermediate prediction layer between the high-level task and the low-level action. \acro trains to map tasks described in language into \skills, and then uses the inferred \skill to predict the action, where both steps are conditioned on visual input and the task. We label \skills using an automated procedure that scales to a wide variety of tasks at no human labeling cost. We instantiate \acro using a single transformer model like RT-2~\cite{rt22023arxiv}, where both the action and \skill queries are co-trained with a vast amount of internet-scale data. \acro (1) enables more data sharing between different tasks by learning the shared task structure across seemingly disparate tasks, and thus is more capable of ingesting multi-task datasets at scale, and (2) is amenable to \corrections that change the underlying behaviors within the context of the scene and task. In our experiments, we show that \acro outperforms RT-2 and action hierarchy ablations on diverse multi-task data. Then we show that RT-2 is highly correctable in \skill space even for unseen \skills, and that learning from these \corrections outperforms learning from teleoperation-based corrections. Finally, we show that \acro is more robust to scene and object variations compared to RT-2. These results show the promise of action hierarchies using language, and we believe \acro provides a strong foundation on which to scale up data collection and robot learning.

\smallskip \noindent \textbf{Limitations \& Future Work}: \acro opens several exciting avenues for future work. First, we test \acro on a large and diverse datasets, achieving state-of-the-art performance, but the absolute success rates still leave room for improvement, even after training on corrections. We believe that, as evidenced by the more sample efficient \corrections of \acro, future work should scale up both the offline datasets and correction pipeline -- \skills could even be used to help bridge datasets with many different embodiments like OXE~\cite{open_x_embodiment_rt_x_2023}, or even to learn from human videos with actions described only in language.

Second, although we ablate different action hierarchies in \cref{sec:exp:diverse}, future work is needed determine the best abstraction level for the intermediate layers (e.g., using object-referential language vs. our \skills). Additionally, we primarily test only one intermediate action layer in this work, or just one step of action reasoning, but future work might define multiple steps of action reasoning instead, where the user can intervene at any level of abstraction. For example, we might add a \emph{task} prediction level to go from long horizon instructions like ``clean the room" to individual tasks like "pick coke can", before mapping the task to \skills and then actions. To decompose corrections at any level of the hierarchy, one might even teach the model to automatically \emph{locate} a correction in the hierarchy, and then autoregressively predict lower level actions until it deems it has reached the robot action level.

Third, \skills represent a contextual and compressed space in which to predict actions. One might leverage this motion contextuality in \acro to greatly compress the action space for reinforcement learning methods and policy exploration, possibly leveraging \skill prediction as a prior. We suspect \skills will provide more meaningful degrees of exploration and more sample efficient policy learning, while also being highly interpretable and correctable to humans. Even in imitation learning, several works have shown the important of action consistency across demonstrators~\cite{mandlekar2021what, belkhale2023data}, and we posit that using \skills as a compressed action space could lead to more consistent actions and thus more sample-efficient policy learning.

\section*{Acknowledgments}
We would like to thank Grecia Salazar, Deeksha Manjunath, Clayton Tan, Yansong Pang, Jornell Quimbao, Tran Pham, Utsav Malla, April Zitkovich, and Elio Prado for their help with dataset collection and on robot evaluation. We would also like to thank the greater Google DeepMind team for their feedback and contributions.


\bibliographystyle{unsrtnat}
\bibliography{references}

\cleardoublepage
\appendix
\noindent We first outline the implementation of \acro and ablations in \cref{app:method_details}, along with the training recipes. Then we discuss implementations and training protocols for methods using corrections in \cref{app:interventions}. After, we detail each of the datasets used in \cref{app:datasets}. Then, we show more detailed results in \cref{app:more_results}, including success rates for different stages of each task for each method, \RV{quantitative analysis of contextuality in \acro, qualitative analysis of the \skill multimodality of \acro. Finally we cover some frequently asked questions about \acro.}

\subsection{Method Implementations}
\label{app:method_details}
\noindent As described in \cref{sec:method}, \acro and RT-2~\cite{rt22023arxiv} are implemented using a Pali-X 55B Multimodal Encoder Decoder Transformer architecture~\cite{chen2023palix}. Images are encoded using a 22B ViT architecture, which is learned during the Pali-X pre-training phase but fixed during robot demonstration data co-training. The encoded images and the prompt are passed through the Encoder, and the output for each query is autoregressively decoded with the Decoder. Next we describe each method (offline only) in detail.

\smallskip
\noindent \textbf{\acro}: \acro first predicts \skills from the task using the following \textbf{\skill query}: \emph{Q: What skill should the robot do to [task]? A: }, where the task is specified in language, and the resulting \skill (skill) is returned in language.
Then, it uses the predicted \skill to better inform action prediction using the following \textbf{action query}: \emph{Q: What action should the robot do to [task], with current skill: [motion]? A: }, where the output is the tokenized action string.
\acro trains in the same fashion as RT-2~\cite{rt22023arxiv}. First, we pretrain \acro on the same large vision language dataset as RT-2, and we then co-train the \skill and action queries using 50\% of the overall training samples (25\% each), along with 50\% using the original pre-training mixture. Similar to RT-2, we use a learning rate of 1e-3 and with a constant linear warmup and square root normalize decay, and a batch size of 1024.

\noindent \textbf{\joint}: Unlike \acro, \joint uses a single query to predict both \skill and action. While both methods are autoregressive on the \skill, \acro has two queries which using different wordings to indicate \skill or action prediction, and \acro also passes in the \skill to the encoder for the action query since it is part of the prompt string. The prompt for \joint is as follows: \emph{Q: What skill and action should the robot do to [task]? A: }. Then the output is a concatenation of first \skill (skill) in language, and then the tokenized action string, in the form \emph{skill: [skill], action: [action]}.
\joint is trained identically to RT-2 and \acro as well, but with the joint \skill and action query instead of action prediction from RT-2.

\smallskip
\noindent \textbf{\cluster}: \cluster follows the same two query procedure and training implementation as \acro. In order to determine the action clusters, we first normalize the actions in the dataset using dataset statistics. Then, we cluster the actions using K-means~\cite{lloyd1982kmeans} using 256 cluster centers. We chose this number to be on par with the number of actively used \skills in the dataset from our automated labeling procedure. Then, the cluster centers are replaced with integers from 0 to 255, and used in place of \skills in the action hierarchy. This ablation tests the utility of \skills compared to embeddings tuned to the specific actions in the datasets.

\smallskip
\noindent \textbf{\onehot}: \onehot also follows the same two query procedure and training implementation as \acro. The only change is to replace unique \skills with integers. We first enumerate skills in order of how common they are, and then assign a unique integer value to each. Importantly, this formulation does not capture the inherent structure of language: for example, ``move arm forward" and ``move arm forward and left" are similar in many ways and should be treated as such, but their replaced one-hot labels will likely be as equidistant as any other two random \skills. Thus, \onehot tests the importance of the structure of language when predicting \skills.

\subsection{Corrections}
\label{app:interventions}

\noindent As described in \cref{sec:method:intervention}, \acro enables humans to intervene with new \skills, and then these corrections can be deployed on robot. To train \acro on corrections, we can directly type or say \corrections that will be passed directly into the action query in place of the inferred \skill from the \skill query. This shifts the burden of correction up one level in the action hierarchy, from actions to \skills. We collect the dataset of \corrections, recording the observations, task, and \corrections, and then co-train our model with the original pre-training dataset, the robot demonstration dataset, and upweighted \corrections. For such a large demonstration dataset, we aim for each correction sample to be seen 50x as often as a corresponding demonstration dataset example. Thus the sampling weights during training on the \diverse dataset are as follows:
\begin{itemize}
    \item Pre-training Queries: 50\%
    \item Demonstration Data \skill Query: 23\%
    \item Demonstration Data Action Query: 23\%
    \item Correction Data \skill Query: 4\%
\end{itemize}
Given that the ratio of the demonstration dataset size to the \correction dataset size is roughly 300:1, this corresponds to upweighting each \correction sample by 50:1.

We use the same recipe for training from teleoperated corrections with RT-2-IWR. The only difference is that the \skill training queries are replaced with action queries like in IWR.

\subsection{Datasets}
\label{app:datasets}

\noindent We use two datasets in this work, the \rt dataset from RT-1~\cite{brohan2022rt1} and RT-2~\cite{rt22023arxiv}, and the new \diverse dataset (which is an extended version of \rt).  \rt consists of the \textbf{6 semantic tasks used for evaluation in 70K demonstrations}, across several common object categories like cans, bottles, and fruits, forming 542 unique instructions. The semantic task instructions are as follows:
\begin{itemize}
    \item \textbf{knock over}: Knock \texttt{Object} Over.
    \item \textbf{drawer place}: Pick \texttt{Object} from drawer and place it on counter.
    \item \textbf{move}: Move \texttt{Object} Near \texttt{Object}.
    \item \textbf{pick}: Pick \texttt{Object}.
    \item \textbf{open / close drawer}: [Open / Close] [Top / Middle / Bottom] Drawer.
    \item \textbf{place upright}: Place \texttt{Object} Upright.
\end{itemize}

\noindent \diverse consists of all the demonstrations from \rt, but with \textbf{24 more semantic tasks in only 30K additional demonstrations}, with 165 unique instructions. The new task instructions in the dataset are as follows, sorted by frequency (most to least):
\begin{itemize}
    \item pull napkin out of dispenser and place napkin flat on counter
    \item \textbf{pull napkin out of dispenser}
    \item \textbf{pick a bowl and place the bowl upright on counter}
    \item \textbf{close the large glass jar containing pistachios using the lid on counter}
    \item pick a cup and place the cup upright on counter
    \item \textbf{open the large glass jar with pistachios}
    \item \textbf{grab a scooper}
    \item pick up the scoop from the basket
    \item open the large glass jar with pistachios and place the lid on counter
    \item place the scoop inside the basket
    \item \textbf{put a bowl under the cereal dispenser spout}
    \item \textbf{move the bowl away from underneath the spout}
    \item \textbf{pick an oatmeal packet and place the oatmeal packet in the bowl}
    \item pick up spoon and place spoon in bowl with cereal
    \item swivel the cereal dispenser until the bowl is half full
    \item pick up the tong from the basket
    \item place the tong inside the basket
    \item pour the snack from the scoop into the cup
    \item scoop the snack from the jar
    \item pick \texttt{object}
    \item move \texttt{object} near \texttt{object}
    \item knock \texttt{object} over
    \item place \texttt{object} upright
    \item squeeze honey into the bowl
\end{itemize}

\noindent This represents a diverse range of behaviors for the robot, often with huge data imbalances between tasks. Note that there are additional demonstrations for the knock over, move, pick, and place tasks in this dataset, although these comprise a small fraction of the overall data. The tasks used for evaluation in \cref{sec:exp:diverse} and 
\cref{sec:exp:intervention} are bolded.

\subsection{Detailed Results}
\label{app:more_results}

\begin{figure*}[h]
    \centering
    \includegraphics[width=\textwidth]{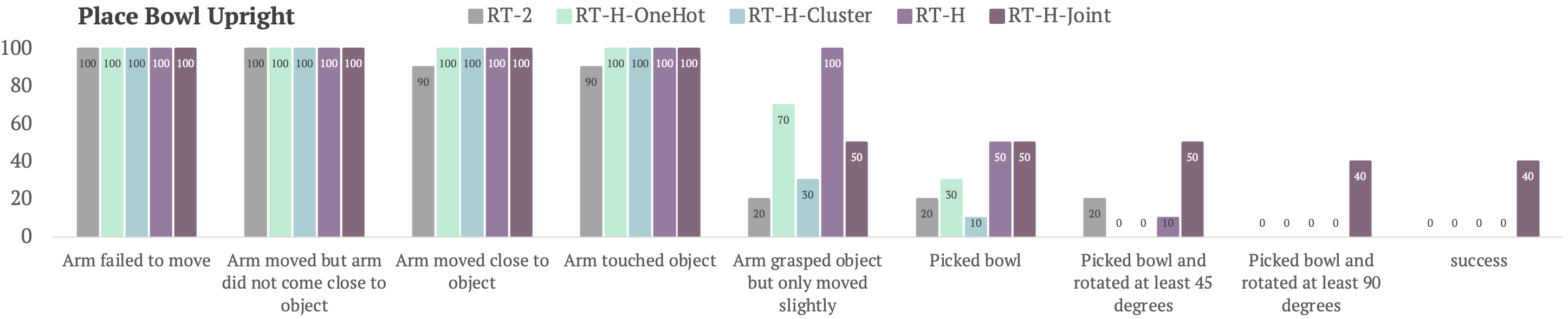}
    \caption{\textbf{Place Bowl Upright on Counter}: Cumulative success rates for each method.}
    \label{fig:bowl_upright}
\end{figure*}

\begin{figure*}[h]
    \centering
    \includegraphics[width=\textwidth]{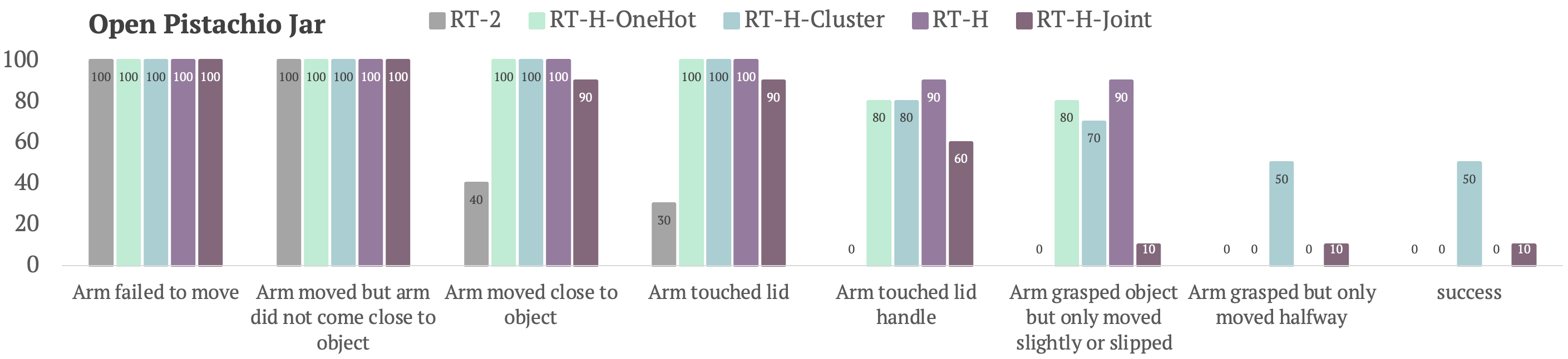}
    \caption{\textbf{Open Pistachio Jar}: Cumulative success rates for each method.}
    \label{fig:open_jar}
\end{figure*}

\begin{figure*}[h]
    \centering
    \includegraphics[width=\textwidth]{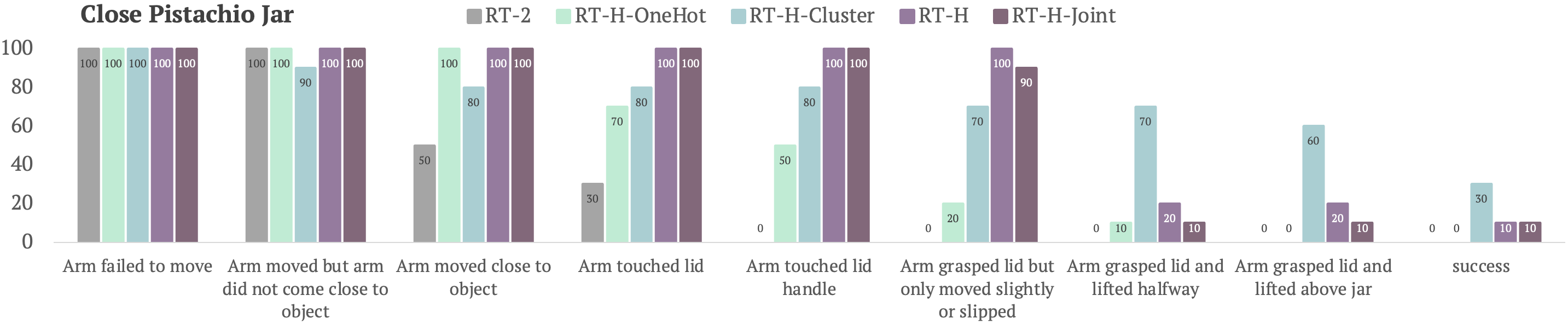}
    \caption{\textbf{Close Pistachio Jar}: Cumulative success rates for each method.}
    \label{fig:close_jar}
\end{figure*}

\begin{figure*}
    \centering
    \includegraphics[width=\textwidth]{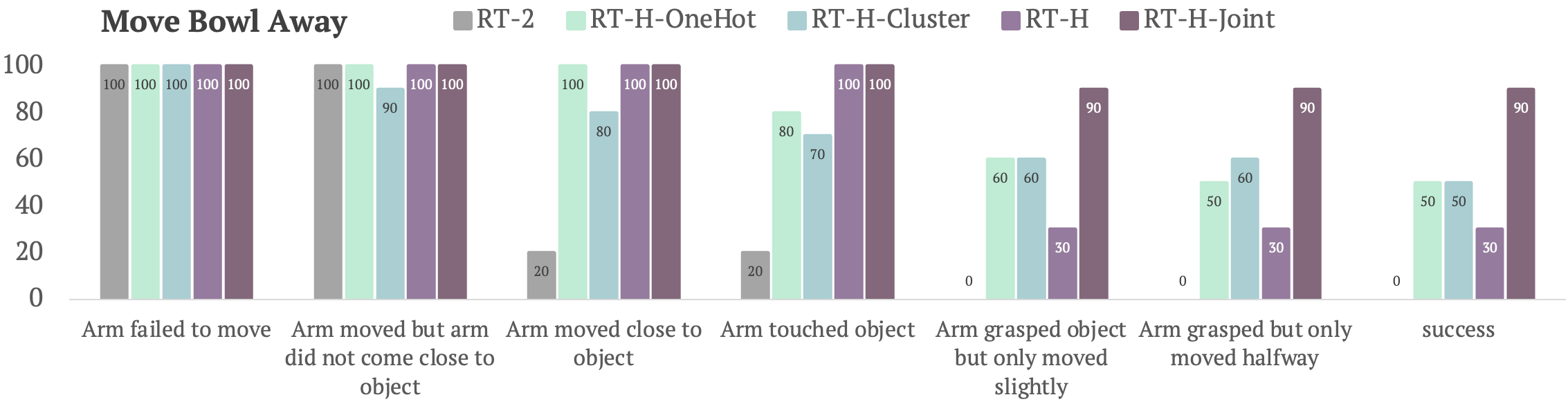}
    \caption{\textbf{Move Bowl Away from Cereal Dispenser}: Cumulative success rates for each method.}
    \label{fig:move_away}
\end{figure*}

\begin{figure*}[h]
    \centering
    \includegraphics[width=\textwidth]{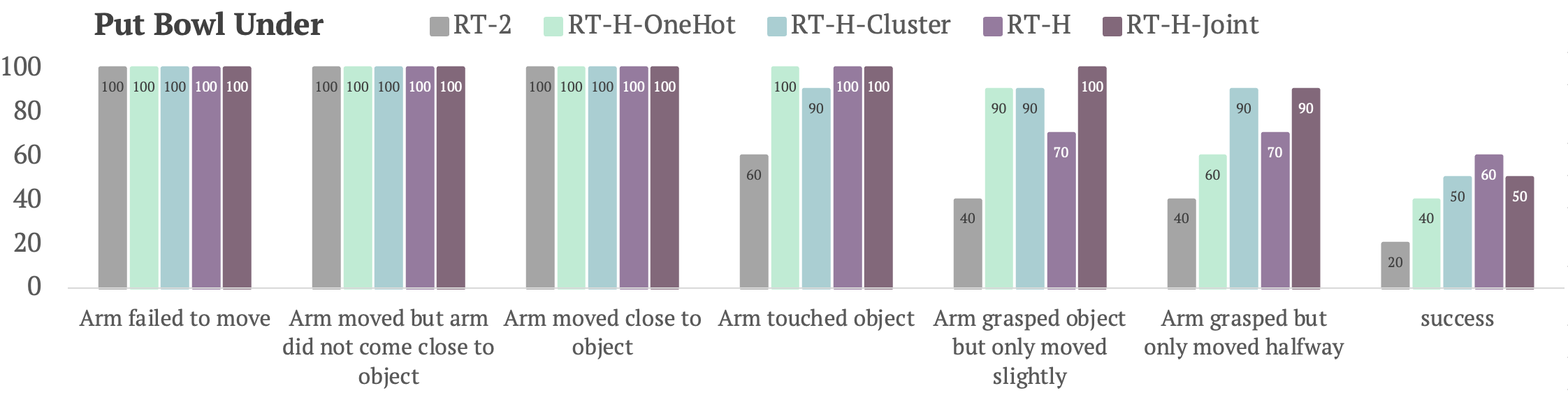}
    \caption{\textbf{Put Bowl Under Cereal Dispenser}: Cumulative success rates for each method.}
    \label{fig:put_under}
\end{figure*}

\begin{figure*}[h]
    \centering
    \includegraphics[width=\textwidth]{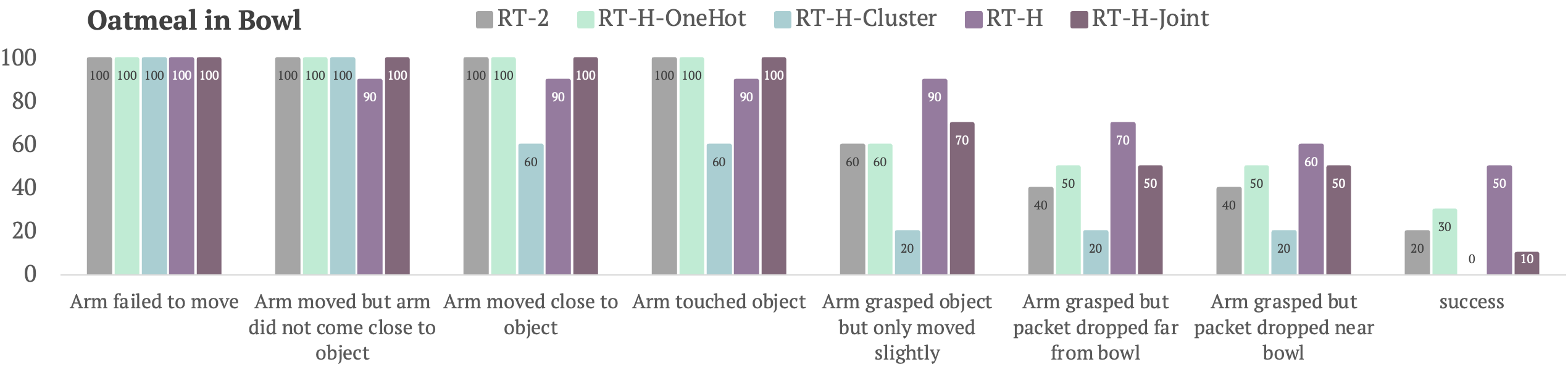}
    \caption{\textbf{Place Oatmeal Packet in Bowl}: Cumulative success rates for each method.}
    \label{fig:oatmeal}
\end{figure*}

\begin{figure*}[h]
    \centering
    \includegraphics[width=\textwidth]{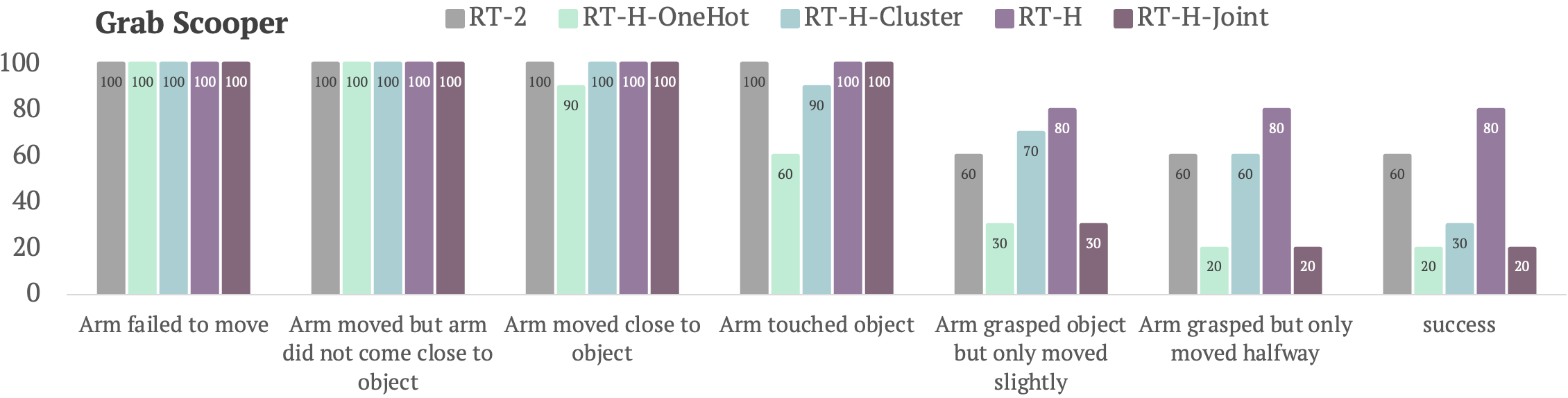}
    \caption{\textbf{Grab a Scooper}: Cumulative success rates for each method.}
    \label{fig:grab_scoop}
\end{figure*}

\begin{figure*}
    \centering
    \includegraphics[width=\textwidth]{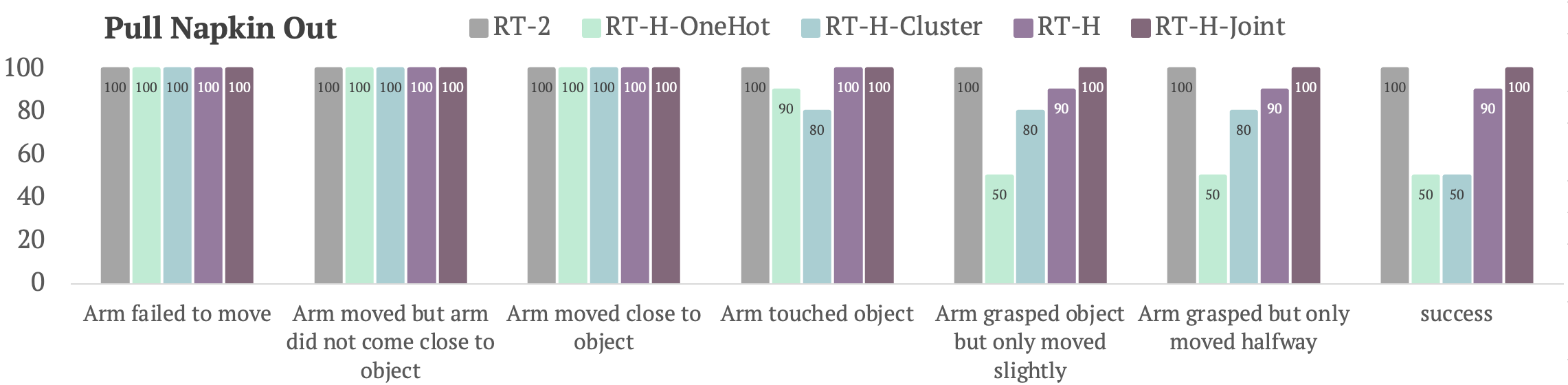}
    \caption{\textbf{Pull Napkin out of Dispenser}: Cumulative success rates for each method.}
    \label{fig:napkin}
\end{figure*}

\noindent \textbf{Diverse Evaluations}: Next we show the cumulative success rates for different stages of each task in the \diverse evaluations from \cref{sec:exp:diverse}. \cref{fig:bowl_upright} shows the ``Place Bowl Upright" task, and \acro and \joint are able to pick up the bowl 50\% of the time (compared to 20\% for RT-2), but \acro struggles to rotate the bowl afterwards. \cref{fig:open_jar} shows the ``Open Pistachio Jar" task, where we see that methods with action hierarchy get substantially farther than RT-2 on this task. \cref{fig:close_jar} shows the ``Close Pistachio Jar" task, where once again RT-2 rarely exhibits the correct behavior compared to methods with action hierarchy. Thus even though success rates for all methods are fairly low on the open and close jar tasks, we see that \acro and its variants are able to progress much farther. \cref{fig:move_away} shows the ``Move Bowl Away" task, where we see once again that methods with action hierarchy get much farther in the task than RT-2. Here, we can see that \acro struggles to grasp the thin rim of the bowl, compared to \joint which has high success with grasping. \cref{fig:put_under} shows the ``Put Bowl Under" task, where once again \acro and other action hierarchy methods do better on each stage of the task, with \acro getting the highest final success rate. \cref{fig:oatmeal} shows the ``Place oatmeal in bowl" task, and \acro and \joint get much farther in the task compared to RT-2, \onehot, and \cref{fig:grab_scoop} shows the ``Grab Scooper" task, and it is one of the few tasks where RT-2 does better than some action hierarchy methods (\cluster and \onehot), but \acro outperforms RT-2 on all stages of the task. In \cref{fig:napkin}, we show the ``Pull napkin out" task, and \acro, \joint, and RT-2 all get very high success rates.

Overall, we see that in many cases, there are only one or two stages of the task that require correction. This often only requires a few language corrections, which provides insight as to why \iv can improve task performance with so little new data.

\begin{figure*}
    \centering
    \includegraphics[width=0.8\textwidth]{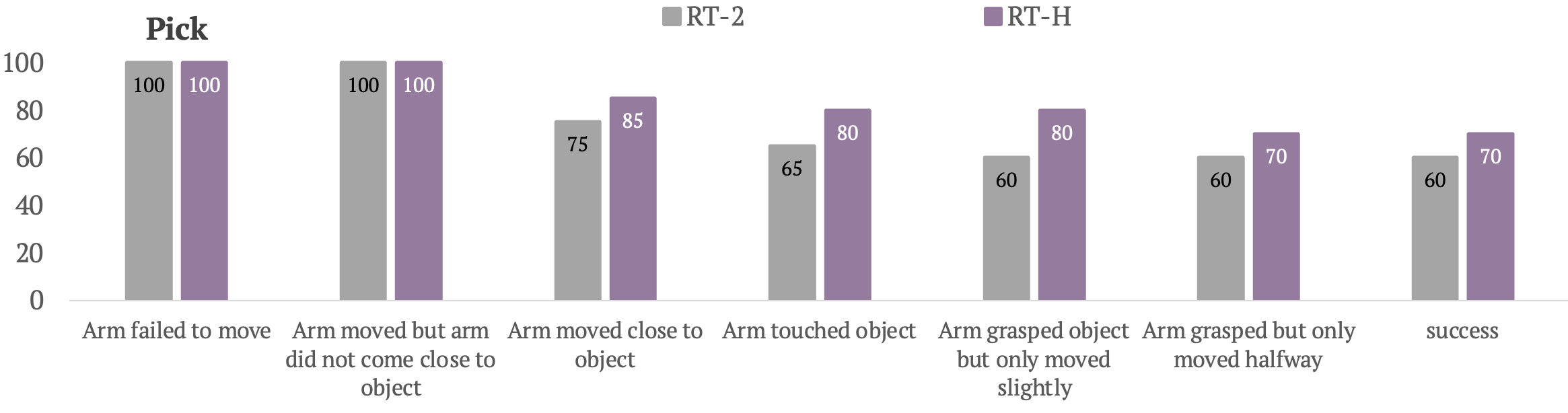}
    \caption{\textbf{Pick (novel objects)}: Cumulative success rates for RT-2 and RT-H. RT-H not only has higher final success rates compared to RT-2, but also success at each stage of the task.}
    \label{fig:pick_novel}
\end{figure*}

\begin{figure*}
    \centering
    \includegraphics[width=0.8\textwidth]{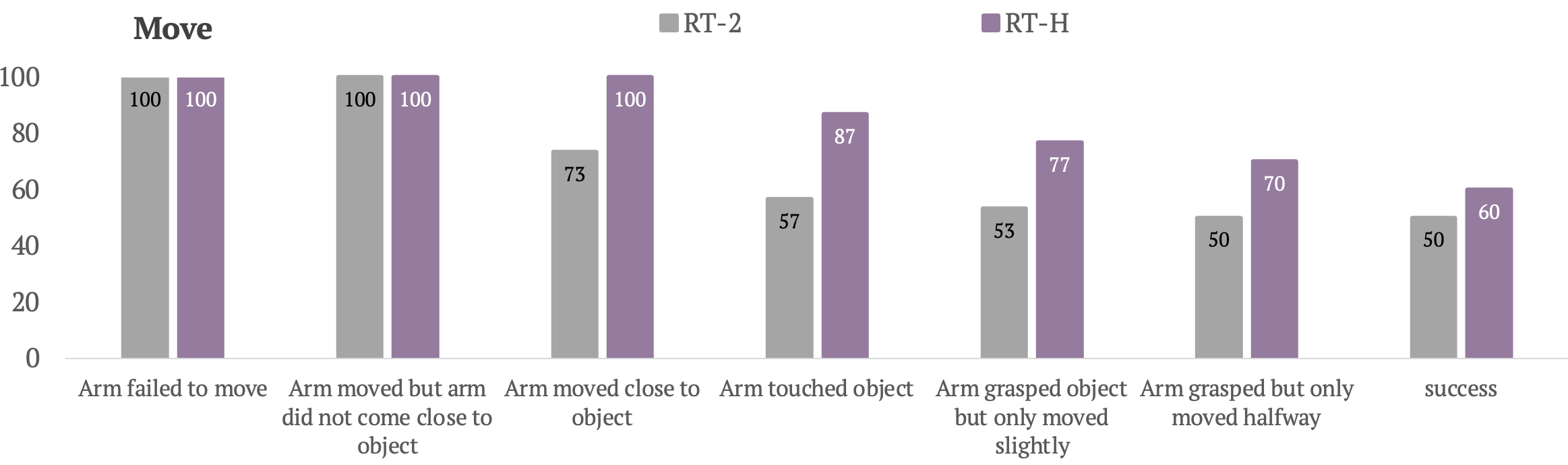}
    \caption{\textbf{Move (novel objects)}: Cumulative success rates for RT-2 and RT-H. RT-H not only has higher final success rates compared to RT-2, but also success at each stage of the task.}
    \label{fig:move_novel}
\end{figure*}

\smallskip \noindent \textbf{Generalization}: Next we show the staged cumulative success rates for \acro generalizing to novel objects, as shown in \cref{sec:exp:generalization} and \cref{tab:generalize_objects}. \cref{fig:pick_novel} shows the pick task and \cref{fig:move_novel} shows the move task, and in both tasks we see that RT-H does better not just in final success rate but also in each individual stage of each task.

\smallskip \noindent \RV{\textbf{Contextuality}: To highlight the contextuality of \acro quantitatively, we compute in \cref{tab:contextuality} the mean (and standard deviation) of each action dimension for actions that belong to the same \skill group. We use the validation set of the \diverse dataset (using the automated \skill labeling procedure from \cref{sec:method:labeling}) to compute these statistics. We find that even though the dominant action dimension for each \skill has the largest mean and action variance (bold in \cref{tab:contextuality}), other action dimensions also have nontrivial variance, suggesting that the interpretation of each \skill changes with the scene and the task. In other words, translating \skill to action (action query) is a contextual process. Sometimes, the mean of the non-dominant action axis is also nontrivial (e.g., for rotate arm right, the arm has some arm (x) and (y) bias), which is likely due to bias from the chosen set of tasks in the dataset.}

\begin{figure*}
    \centering
    \includegraphics[width=\linewidth]{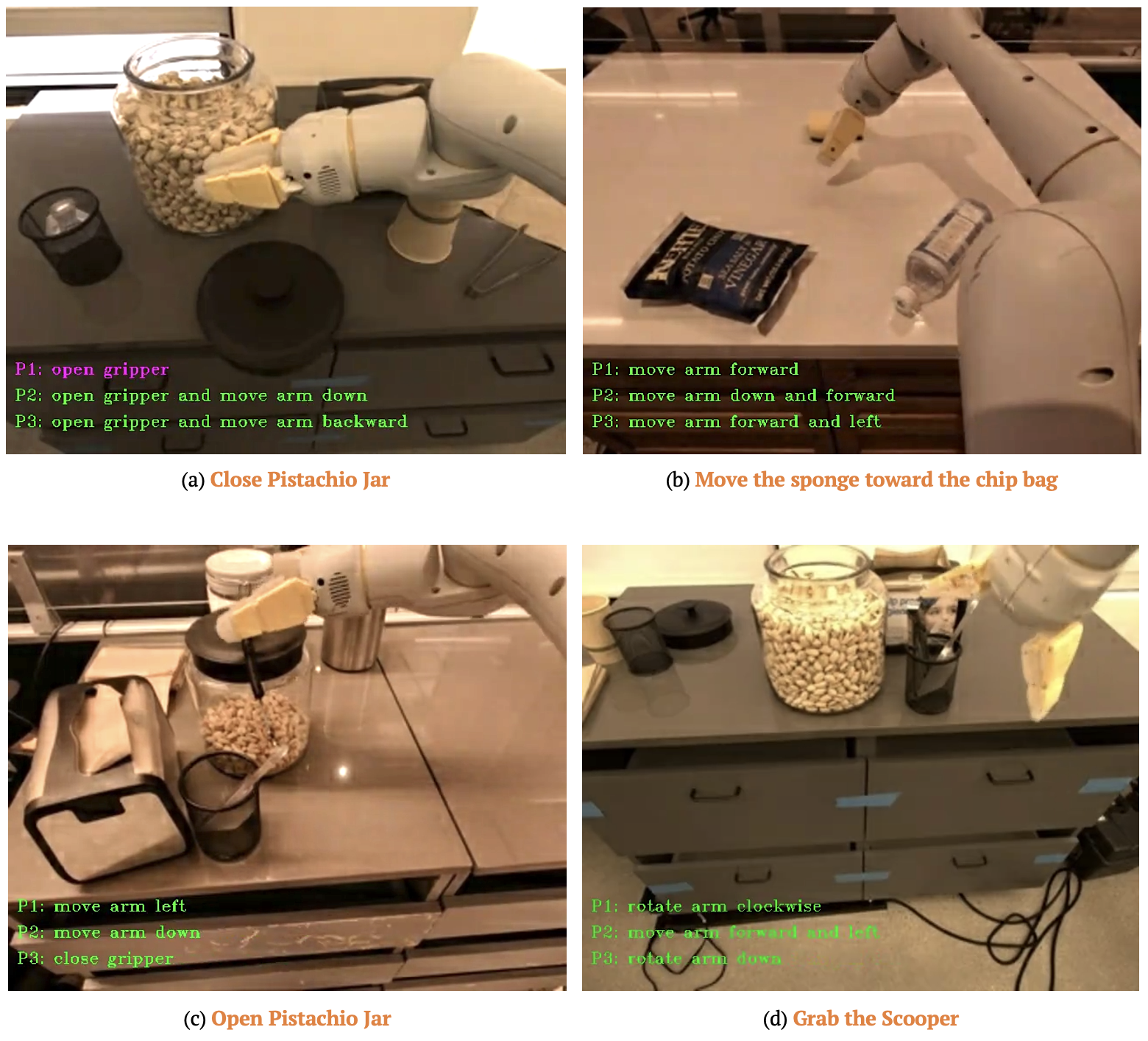}
    \caption{\RV{Four examples of multimodal \skill prediction in \acro using beam search on the \skill query. We show the top three \skill predictions (P1, P2, and P3) for each example. In (a) and (b) (top row), we see that \acro is capable of representing multiple valid ways of doing the task that are all very contextual. The \skills outputted in (a) and (b) are quite similar to each other, but differ in subtle but task-relevant ways. In (c) and (d), we see that \acro predicts even more diverse ways to accomplish the task, once again contextual to the task and scene.}}
    \label{fig:multimodal}
\end{figure*}

\smallskip \noindent \RV{\textbf{Multimodality}: Next, we study if the \skill abstraction has enabled \acro to learn not just the correct \skill at each step, but also diverse ways of accomplishing the same task. To analyze this qualitatively, we run the \skill query on offline validation data with beam search to output the top three \skills for images in the dataset. We show four examples of this in \cref{fig:multimodal}. In the first row (examples (a) and (b)), \acro predicts \skills that differ slightly from each other but in task-contextual ways (e.g., move arm forward vs. move arm down and forward, both are accurate for the task). In the second row (examples (c) and (d)), \acro predicts \skills that are quite different from each other (e.g., move arm left vs. close gripper), but despite the variety, each \skill is reasonable given the context of the scene and task. This shows that \acro can capture the multimodality of behaviors for tasks from the data. In fact, this ability to represent multiple high level behaviors could be how \acro is so efficient in learning from \skill intervention data -- an intervened \skill might be quite likely already in the model, and so updating the model to predict the new \skill might be a trivial change to the model. Additionally, this \skill multimodality could be quite useful for exploration in a reinforcement learning context (e.g., sampling \skills from the model instead of actions from some uniform distribution).}

\subsection{\RV{Frequently Asked Questions}}
\label{appendix:faq}

\smallskip
\noindent
\RV{\textbf{Why does \acro outperform RT-2 conceptually, despite using the same observations and low-level actions?}}

\noindent \RV{Predicting an intermediate action abstraction \emph{structures} the action prediction problem into two stages, each simpler to learn than the combined objective. Learning the relationship between each image, task and each low-level action is a very high-dimensional mapping to learn, especially in multi-task datasets. Language motions provide a bottleneck that reduces that dimensionality for each stage of action prediction. Predicting language motions from the task is much simpler than predicting each action for each dimension directly, and thus leaves less room for overfitting. In addition, language motions learned from multi-task datasets enable the possibility of transfer across the different datasets. For example, we might not expect any transfer between two distinct tasks such as ``picking up a cup" vs ``pouring to a cup", but both tasks would potentially share a common language motion of  “moving forward” enabling transfer of these low-level motions across tasks/datasets, and potentially allowing for more generalization beyond flat models such as RT-2. Finally, predicting actions from language motions and the task is also much simpler, since the language motion narrows down the space of valid actions for the model to predict. This action ``reasoning" through multiple steps is analogous to why LLMs do better at step-by-step reasoning compared to direct prediction.}

\bigskip
\noindent
\RV{\textbf{How does the action hierarchy in \acro compare to LLM planning methods like SayCan~\cite{brohan2023can}?}}

\noindent \RV{The action hierarchies in SayCan and other LLM planning frameworks address a fundamentally different problem: long-horizon instruction following. They often start with a long-horizon \textbf{instruction} (e.g. ``bring me a cold drink") that breaks down to medium horizon \textbf{tasks} (e.g. ``open the fridge" + ``pick up a coke can" + ``close the fridge" + ``bring the coke can to the table") via an LLM or task planner, and the medium horizon task, e.g. ``pick up a coke can" relies on existing pretrained primitives and often an affordance value function to shape the LLM predictions. In \acro, we instead learn action hierarchies from medium horizon \textbf{tasks} (e.g. ``pick up a coke can") to short horizon \textbf{motions} (e.g. ``move arm forward").}

\RV{Therefore, the closest analog to an approach like SayCan in our setting would be using predefined \skill primitives (e.g., hardcoded versions of ``move arm forward" or ``close gripper") We did try to implement this idea, but we found it was impractical for a few reasons:
\begin{enumerate}
    \item There is significant \emph{contextuality} of \skills required when solving precise manipulation tasks (see \cref{fig:context}, e.g., the speed or direction variety for a single \skill) – there was no single predefined primitive for many \skills that could safely and efficiently progress at the task. See \cref{app:more_results} for a quantitative analysis of the contextuality of each major \skill in the dataset. We find that each \skill has nontrivial variation in multiple action dimensions, not just the major action dimension.
    \item LLMs would inherently struggle to predict \skills because they are not grounded in the visual context of the scene. Therefore we would not expect these models to understand directions like ``left" and ``up" or to know when to close the gripper with just a textual description of the scene (as provided in SayCan). Thus VLMs are much better suited for this task.
\end{enumerate}}

\bigskip
\noindent
\RV{\textbf{Can large VLMs be used to directly predict \skills without the need for automated labeling?}}

\noindent \RV{We explored this idea with state of the art VLMs like GPT-4V~\cite{2023GPT4VisionSC}, but were unable to get reasonable \skill outputs. This is likely because current VLMs do not have strong spatial priors on robot behaviors (i.e., they are not grounded well in concepts like ``left" or ``forward"). More recent work like PIVOT~\cite{nasiriany2024pivot} show that VLMs do better with visual spatial information (i.e., arrows drawn on the image), but there are still long term questions about how to incorporate rotations or gripper actions under these frameworks. We show that our automated labeling procedure provides a robust and scalable way to teach VLMs like RT-H these spatial concepts. As VLMs gain more and more knowledge of the physical world, we are excited to see how they improve at predicting \skills zero-shot.}

\bigskip
\noindent
\RV{\textbf{Have we considered scenarios where the task involves tool use, and how \skills can evolve to describe them?}}

\noindent \RV{As we noted in the future work section, we hope to expand \skills to include object-referential language like ``rotate the screw" or ``grasp the pot handle". We believe this will unlock a whole new set of capabilities and types of corrections. However the main challenge is how to get these types of \skills without expensive human annotation. One idea is to label a fraction of the data with object-referential language, and label the rest of the data using a model trained on this dataset or through self-supervised techniques. As VLMs become more powerful, it might even be easier for VLMs to zero-shot provide these object-referential \skills than the spatial \skills we use in \acro.}

\bigskip
\noindent
\RV{\textbf{Does co-training on intervention data lead to model collapse, where the model worsens at tasks not present in the intervention data?}}

\noindent \RV{We did not notice any model collapse in practice – specifically, we noticed that offline metrics on different splits of the data remained stable after adding in the intervention data. Likely due to the immense parameter counts, \acro and other large VLMs seem very capable of integrating new data under co-training schemes. }

\bigskip
\noindent
\RV{\textbf{Does the asynchronous \skill inference in \acro hurt policy performance?}}

\noindent \RV{Inference time greatly increases for synchronous inference, and unfortunately this can make online evaluation of the synchronous procedure quite difficult. Instead, we can turn to the offline metrics in \cref{tab:mse}, where we see \acro has a lower MSE (this uses the asynchronous inference procedure but offline) than \joint and RT-2. As we note in the paper, the comparison between \acro and \joint can also be seen as a comparison between asynchronous and synchronous inference, and there we see fairly minor performance differences between the two (if any).}

\begin{table*}[h]
    \centering
    \begin{tabular}{c||ccc|ccc|c}
        \skill & arm (x) &	arm (y) &	arm (z) &	arm (rx) &	arm (ry) &	arm (rz) &	gripper \\ \hline 
        \emph{move arm forward} &	\textbf{\MS{7.7}{6.2}} &	\MS{-0.5}{3.3} &	\MS{-1.4}{3.2} &	\MS{3.4}{10.9} &	\MS{-1.1}{8.9} &	\MS{-4.0}{10.3}	& \MS{1.0}{3.3} \\
        \emph{move arm backward} &	\textbf{\MS{-10.9}{10.1}} &	\MS{-2.4}{4.6} &	\MS{-2.3}{5.2} &	\MS{0.5}{9.2} &	\MS{7.8}{11.6} &	\MS{0.0}{13.1}	& \MS{0.5}{6.6} \\
        \emph{move arm left} &	\MS{-0.4}{3.2} &	\textbf{\MS{8.2}{6.3}} &	\MS{-2.1}{3.4} &	\MS{6.1}{12.0} &	\MS{2.5}{7.7} &	\MS{7.9}{11.5}	& \MS{1.0}{2.1} \\
        \emph{move arm right} &	\MS{-1.4}{4.1} &	\textbf{\MS{-6.9}{7.3}} &	\MS{-1.5}{4.0} &	\MS{-1.9}{11.7} &	\MS{2.2}{7.7} &	\MS{-6.6}{11.6}	& \MS{1.1}{0.9} \\
        \emph{move arm up} &	\MS{-1.7}{4.6} &	\MS{-2.1}{4.9} &	\textbf{\MS{11.1}{8.6}} &	\MS{1.7}{11.7} &	\MS{-7.4}{10.9} &	\MS{0.5}{10.5}	& \MS{0.9}{4.6} \\
        \emph{move arm down} &	\MS{-0.6}{4.0} &	\MS{-0.3}{3.1} &	\textbf{\MS{-8.1}{9.0}} &	\MS{1.4}{10.0} &	\MS{5.6}{10.5} &	\MS{-1.0}{8.2}	& \MS{1.1}{3.5} \\
        \emph{rotate arm right} &	\MS{2.1}{4.6} &	\MS{3.0}{5.5} &	\MS{0.3}{5.1} &	\textbf{\MS{29.3}{24.5}} &	\MS{-2.1}{12.2} &	\MS{0.0}{13.0}	& \MS{1.0}{1.5} \\
        \emph{rotate arm left} &	\MS{0.4}{4.2} &	\MS{-0.4}{4.2} &	\MS{-0.5}{3.9} &	\textbf{\MS{-24.3}{21.9}} &	\MS{-4.1}{14.8} &	\MS{0.9}{10.6}	& \MS{1.2}{1.6} \\
        \emph{rotate arm up} &	\MS{0.0}{4.5} &	\MS{0.3}{3.3} &	\MS{-2.3}{4.4} &	\MS{2.5}{12.4} &	\textbf{\MS{20.5}{18.2}} &	\MS{1.7}{8.8}	& \MS{1.0}{1.4} \\
        \emph{rotate arm down} &	\MS{0.3}{4.5} &	\MS{0.8}{3.8} &	\MS{1.9}{5.0} &	\MS{-7.4}{15.1} &	\textbf{\MS{-28.7}{26.6}} &	\MS{-0.1}{13.6}	& \MS{1.0}{1.6} \\
        \emph{rotate arm counterclockwise} &	\MS{3.3}{4.3} &	\MS{-1.1}{4.8} &	\MS{-0.3}{4.5} &	\MS{4.2}{12.7} &	\MS{-3.1}{12.0} &	\textbf{\MS{-24.1}{21.5}}	& \MS{1.1}{3.4} \\
        \emph{rotate arm clockwise} &	\MS{-0.5}{5.4} &	\MS{3.1}{5.1} &	\MS{0.0}{4.8} &	\MS{-0.9}{12.3} &	\MS{-0.1}{11.8} &	\textbf{\MS{24.0}{20.8}}	& \MS{0.9}{5.2} \\
        \emph{open gripper} &	\MS{0.6}{0.9} &	\MS{0.7}{1.3} &	\MS{0.7}{1.3} &	\MS{0.7}{2.7} &	\MS{0.6}{2.0} &	\MS{0.8}{2.3}	& \textbf{\MS{-67.6}{42.9}} \\
        \emph{close gripper} &	\MS{0.7}{1.8} &	\MS{0.8}{1.3} &	\MS{0.8}{2.0} &	\MS{0.9}{4.3} &	\MS{0.5}{3.8} &	\MS{0.9}{3.6}	& \textbf{\MS{65.0}{43.5}} \\
    \end{tabular}
    \caption{\RV{Action Means (and Standard Deviations) for basic \skills (cardinal directions) for each action dimension (arm delta x,y,z; rotate arm delta x,y,z; and gripper) in the \diverse dataset, computed over the validation set. The bolded numbers correspond to the dominant axis which the \skill refers to. Note that positions and rotations are not scaled to match each other. We find that while the dominant axis has the largest mean and variance for each skill (bolded numbers), other axes also have nontrivial variation (but often close to zero mean). This demonstrates that a given skill is not merely a fixed primitive, but maps to a wide variety of potential actions depending on the actual state and the task (i.e., context).}}
    \label{tab:contextuality}
\end{table*}

\end{document}